\documentclass{article}
\usepackage{arxiv}
\usepackage[english]{babel}
\usepackage{csquotes}
\usepackage[utf8]{inputenc} % allow utf-8 input
\usepackage[T1]{fontenc}    % use 8-bit T1 fonts
\usepackage[colorlinks=true, allcolors=blue]{hyperref}
\newcommand{\email}[1]{\href{mailto:#1}{\tt{\nolinkurl{#1}}}}
\newcommand{\orcid}[1]{ORCID: \href{https://orcid.org/#1}{\tt{\nolinkurl{#1}}}}
\usepackage{authblk}
\usepackage{natbib}
\usepackage{amsmath}
\usepackage{graphicx}
\usepackage{subfig}
\usepackage{float}
\usepackage[colorlinks=true, allcolors=blue]{hyperref}
\usepackage{array}
\usepackage{tabularx} % 加载tabularx宏包
\usepackage[table]{xcolor} % 使用xcolor宏包,带table选项
\usepackage{multirow} % 引入multirow宏包

\usepackage{amsmath,amsfonts,amssymb,mathtools,amsthm}

\newtheorem*{theorem*}{Theorem}

\newtheorem*{lemma*}{Lemma}

\newtheorem*{property*}{Property}

\newtheorem*{assumption*}{Assumption}
\newtheorem*{prop*}{Proposition}

\newtheorem*{setting*}{Setting}

\title{Reasoning Bias of Next Token Prediction Training}

\author[1,2]{Pengxiao Lin}
\author[1,2,*]{Zhongwang Zhang}
\author[1,2,3,4,5,*]{Zhi-Qin John Xu}
\affil[1]{Institute of Natural Sciences, MOE-LSC, Shanghai Jiao Tong University}
\affil[2]{School of Mathematical Sciences, Shanghai Jiao Tong University}
\affil[3]{ School of Artificial Intelligence, Shanghai Jiao Tong University}
\affil[4]{Key Laboratory of Marine Intelligent Equipment and System, Ministry of Education, P.R. China}
\affil[5]{Center for LLM, Institute for Advanced Algorithms Research, Shanghai}
\affil[*]{Corresponding author: \email{xuzhiqin@sjtu.edu.cn}}
\begin{document}
\maketitle
\begin{abstract}
Since the inception of Large Language Models (LLMs), the quest to efficiently train them for superior reasoning capabilities has been a pivotal challenge. The dominant training paradigm for LLMs is based on next token prediction (NTP). Alternative methodologies, called Critical Token Prediction (CTP), focused exclusively on specific critical tokens (such as the answer in Q\&A dataset), aiming to reduce the overfitting of extraneous information and noise. Contrary to initial assumptions, our research reveals that despite NTP's exposure to noise during training, it surpasses CTP in reasoning ability. We attribute this counterintuitive outcome to the regularizing influence of noise on the training dynamics. Our empirical analysis shows that NTP-trained models exhibit enhanced generalization and robustness across various benchmark reasoning datasets, demonstrating greater resilience to perturbations and achieving flatter loss minima. These findings illuminate that NTP is instrumental in fostering reasoning abilities during pretraining, whereas CTP is more effective for finetuning, thereby enriching our comprehension of optimal training strategies in LLM development.
\end{abstract}

\section{Introduction}

As transformer-based Large Language Models (LLMs) continue to fuel enthusiasm for Artificial General Intelligence (AGI), numerous techniques are emerging to advance this trend, fostering a highly optimistic outlook for the eventual realization of AGI. A central challenge since the inception of LLMs has been how to efficiently train these models to achieve superior reasoning capabilities. Over time, a series of training techniques have revolutionized the performance of LLMs, each contributing to significant milestones in the field.

The success of natural language processing (NLP) has been significantly driven by the widespread adoption of next token prediction (NTP), a self-supervised learning approach popularized by the GPT series \citep{GPT1, GPT2, GPT3}. Unlike supervised methods that depend on costly labeled data, NTP enables models to learn from vast amounts of unlabeled text by predicting subsequent tokens, allowing for zero-shot generalization and eliminating the need for task-specific finetuning. This framework has established NTP as a cornerstone of modern NLP.

\begin{figure}[!ht]
    \centering
    \includegraphics[width=0.6\columnwidth]{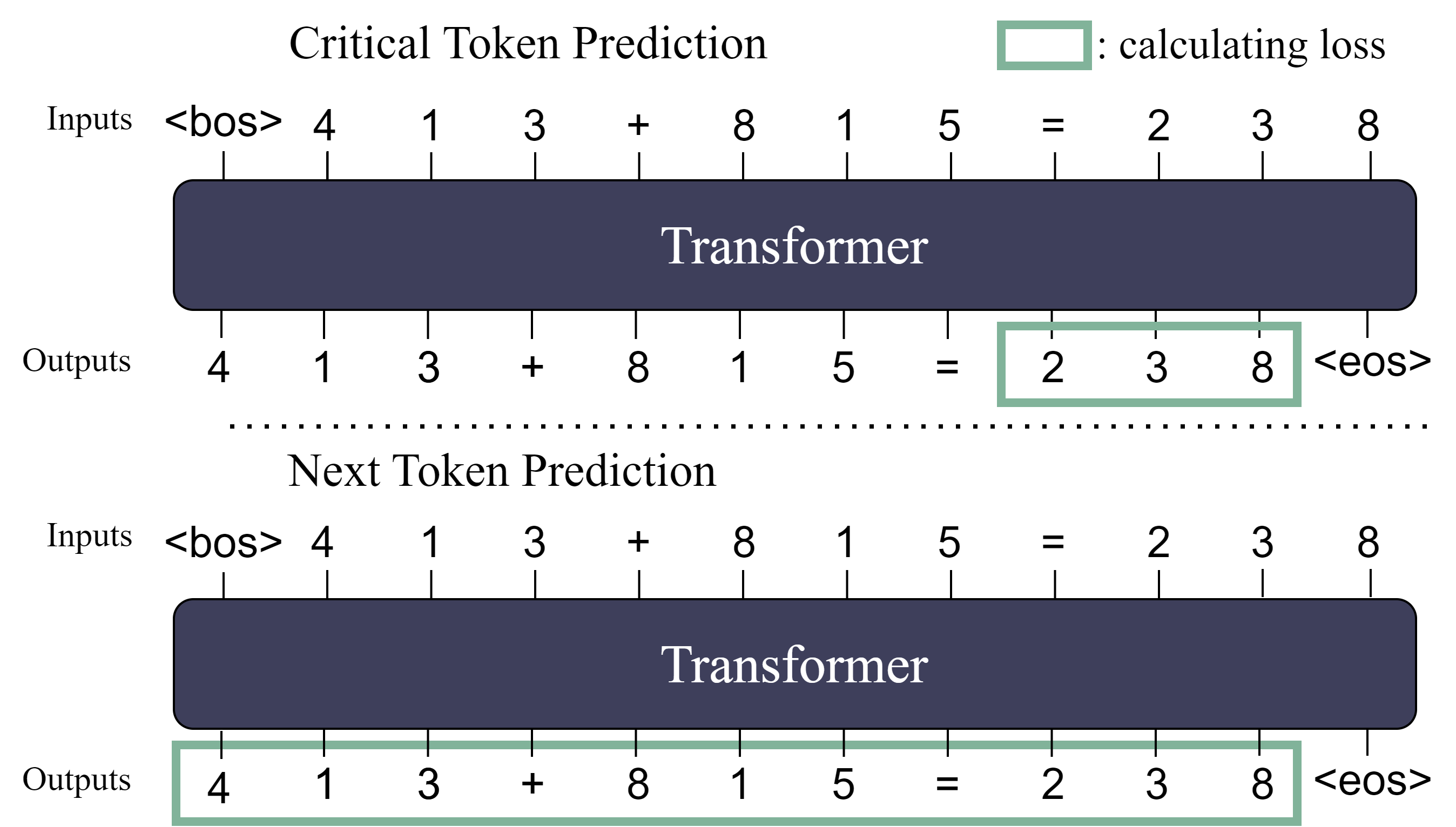}
    \caption{The schematic illustration comparing NTP and CTP. In the context of arithmetic addition tasks, CTP's loss function exclusively focuses on the answer, whereas NTP's loss encompasses the entire sequence, consequently introducing a certain degree of noise during the optimization process.}
    \label{fig:add_flowchart}
\end{figure}

In contrast to NTP, supervised training only on labels can be regarded as critical token prediction (CTP) illustrated in Fig.~\ref{fig:add_flowchart}. Although NTP has been successfully applied in LLMs, it still leaves room for speculation: Given the availability of labeled data, should CTP be reconsidered as a viable alternative? For instance, in training a model for arithmetic addition, employing NTP to learn problem formulations seems inherently flawed, as the subsequent components cannot and should not be inferred from preceding ones in math problems. Furthermore, recent advancements have increasingly focused on the strategic selection of important tokens for training. For example, RHO-1 \citep{rho1} utilizes a model to score each token and trains only on samples with high scores. Phi-4 \citep{abdin2024phi4technicalreport} has made significant strides in enhancing reasoning capabilities through an emphasis on data quality. One key technique involves synthesizing a large number of question-and-answer (Q\&A) data pairs, even during the pretraining phase with NTP. This raises a natural question: since the answer portion of Q\&A data can be seen as a form of label, should CTP be used for Q\&A pairs instead?

In this study, we conduct a systematic comparison between NTP and CTP using academically sound research methods. Our findings reveal that NTP training exhibits superior generalization capabilities compared to CTP across various benchmark datasets focused on reasoning. This suggests that NTP possesses an inherent reasoning bias. To further investigate this bias, we employ PrOntoQA and a form of synthetic data known as the anchor function, providing additional empirical evidence. Beyond the reasoning bias, we observe that models trained with NTP exhibit better transfer capability and demonstrate greater robustness than those trained with CTP when subjected to perturbations in weights or hidden features. 

Thus, during the pretraining phase, NTP can still enhance reasoning performance and robustness over CTP, even for Q\&A pairs. However, in the supervised finetuning stage, NTP shows almost no advantage over CTP. This is likely because the pretraining phase has already guided the training process to the vicinity of a specific minimum. Moreover, as NTP needs to accommodate more noise, its training speed is slower than that of CTP. Consequently, in the supervised training stage, CTP emerges as a more suitable option.

% NTP training also mirrors the language learning process of a child. During the infancy stage, a child will not only repeat the answer part of a Q\&A dialog pair, but also repeat everything they hear. The noise introduced in such NTP learning has a positive effect on generalization, similar to the noise in SGD or Dropout training. Therefore, further investigation into the implicit bias of NTP is an important research direction for understanding LLMs.

\section{Related Work}
\paragraph{Next-Token Prediction and Other Training Methods.}
% Next token prediction is a classic and widely adopted method for training LLMs, enabling them to acquire fundamental expressive capabilities. 
Recent studies have deepened our understanding of NTP through various perspectives. For instance, \citep{zhao2024implicit, thrampoulidisimplicit} analyze the geometric properties of word and context embeddings in the logits domain, revealing the mechanisms behind the sparsity and low-rank structures in logits space. Theoretical explorations by \citep{madden2024nexttokenpredictioncapacitygeneral} further investigate the capacity of NTP in single-layer transformers, focusing on the interplay between model parameters and output dimensions. Additionally, \citep{MechanicsofNextTokenPredictionwithSelf-Attention} leverage knowledge graphs to provide mechanistic insights into the NTP learning strategy, while \citep{heLawNextTokenPrediction2024} establish empirical scaling laws for NTP across diverse language models. Despite its widespread use, concerns about the limitations of NTP have spurred the development of alternative training paradigms. Recent work by \citep{pmlr-v235-bachmann24a, gloeckle2024better} highlights the potential of novel learning methods to address these limitations. For example, RHO-1 \citep{rho1} introduces a token-level scoring mechanism, selectively training on high-scoring samples to improve efficiency. Similarly, Phi-4 \citep{abdin2024phi4technicalreport} demonstrates significant advancements in reasoning capabilities by prioritizing high-quality data during training. While these innovations mark important progress, the relationship between different training methodologies and their corresponding generalization capabilities remains underexplored. A deeper understanding of this relationship is crucial for advancing the field and developing more robust and efficient LLMs.

% \paragraph{Initial Bias in Transformers}
% Regularization is an unavoidable issue in training neural networks. In recent years, the randomness introduced by Stochastic Gradient Descent (SGD) has been shown to lead to flatter solutions with better generalization, a result confirmed both empirically and theoretically \citep{smith2021on, barrett2022implicitgradientregularization, pmlr-v202-andriushchenko23b}. Beyond training implicit regularization, the initial bias introduced in \citep{InitialGuessingBias} significantly influences the features learned by neural networks. The concept of neural redshift, proposed by \citep{Teney_2024_CVPR}, demonstrates that the activation function, layer normalization, and depth of the Transformer largely determine the initial complexity of a network, thereby influencing the frequency of functions the network learns. Furthermore, \citep{Position} find that pretrained LLMs generate lower complexity sequences compared to untrained LLMs, highlighting the critical role of the NTP training method.

\paragraph{\textbf{Implicit Bias for Noise-Induced Regularization Techniques.}}
Implicit bias introduced by noise-induced regularization techniques has been widely studied in recent years. Different forms of noise often have a significant impact on the training process and the final performance of the model~\citep{zhu2019anisotropic}. Among the noise-induced regularization techniques, stochastic gradient descent (SGD) is the most widely studied. A series of works have shown that the noise introduced by SGD can improve the generalization ability of models by making the loss landscape flatter \citep{wu2020noisy,feng2021inverse,xie2020diffusion}. Specifically, \citep{mori2021power} highlight that the magnitude of SGD noise depends on the loss landscape, which is crucial for helping SGD converge to flatter minima. An alternative line of research \citep{wu2018sgd,ma2021linear} links SGD's preference for flatter minima to the dynamical stability of minima.
Dropout is another widely used technique to improve model generalization~\citep{dropdim, dropattention, drophead, li2023dropkey, demand_dropout, UniDrop, he2024matterstransformersattentionneeded}. A series of studies have shown that the noise introduced by dropout can enhance generalization ability from different perspectives~\citep{mianjy2018implicit, bank2020etf, lengerich2022dropout, cavazza2018dropout, wei2020implicit, zhang2023stochastic}. \citet{zhang2024implicit} find the noise introduced by dropout can foster model condensation and improve the flatness of the loss landscape, explaining the reasons for dropout’s improvement of model generalization from two aspects. In this work, we draw an analogy between NTP and noise-induced training methods, and explore the impact of NTP on the model's reasoning capabilities.

\section{Preliminaries}

In this section, we introduce some key definitions of the model architecture and the training methods, next token prediction, and critical token prediction. 

\subsection{Model architecture}
We use the original GPT2-125M structure, which is composed of embedding, transformer block, and projection. Each transformer block contains the self-attention block and the fully connected layer block. For self-attention block $\mathrm{Attn}$ we have
\begin{equation}
            \mathrm{Attn}(X) = \mathrm{softmax}\left(\frac{XW_QW_K^{T}X^T}{\sqrt{d_k}}\right)XW_V.
\end{equation}
And the fully connected block is
\begin{equation}
    \mathrm{MLP}(X) = \mathrm{ReLU}(XW_1)W_2.
\end{equation}
Both prenorm and postnorm settings apply to the conclusions of this paper.
% , which are
% \begin{equation}
%     \begin{aligned}
%         \text{(prenorm)}&\ X_{k+1} = X_k + F_k(\text{Layernorm}(X_k)),\\ 
%         \text{(postnorm)}&\ X_{k+1} = \text{Layernorm}(X_k + F_k(X_k)),
%     \end{aligned}
% \end{equation}
% and the $F_k = \mathrm{Attn}_k$ or $\mathrm{MLP}_k$.
For realistic reasoning tasks, we initialize the weight with zero-mean normal distribution with a standard deviation of 0.02 default by Hugging Face. In the anchor function task, we use the kaiming initialization.  

\subsection{Definition of NTP and CTP}
We note the input sequence with length $T$ in the token format $\{x_k\}_{k=1}^T$, and without loss of generality, the critical token is set as the end token $x_T$. We also denote $P(x_{t+1}|x_{\leq t}) = P(x_{t+1}|x_1, \ldots, x_{t})$ as the model output logits at the $t$-token. Training loss of NTP and CTP are defined as follows:
\begin{align}
    \mathcal L_N &= -\frac{1}{T}\sum_{t=1}^{T-1} \text 1\{x_{t+1}\}\log(P(x_{t+1}|x_{\leq t})), \\
    \mathcal L_C &= -\text 1\{x_{T}\}\log(P(x_{T}|x_{\leq T-1})).
    \label{eq:loss}
\end{align}
Not hard to see the CTP loss $\mathcal L_C$ is the critical part of NTP loss $\mathcal L_N$, only calculated on the critical token $x_T$. The distinction between CTP and supervised finetuning (SFT) lies in model states and task objectives. Unlike SFT, which leverages pretrained models to acquire downstream task-solving capabilities, CTP primarily minimizes interference from textual noise during the pre-training phase.

%Although NTP is a crucial component in the pre-training phase of large language models, there remains a significant gap in the comprehensive understanding of pre-trained models. We initially propose that NTP facilitates transformer-based models in learning more generalized representations, even for QA datasets where the data structure inherently supports CTP training.

% The transformer is set as 8 layers and 8 heads, with 200 hidden state dimension and 64 dimension for $Q, K, V$. We choose ReLU as the activation in feed-forward blocks with dimension 1200. We apply the kaiming normal initialization on the transformer layers.

\section{Experimental results}

\begin{figure*}[!htb]
        \centering
        \includegraphics[width=\textwidth]{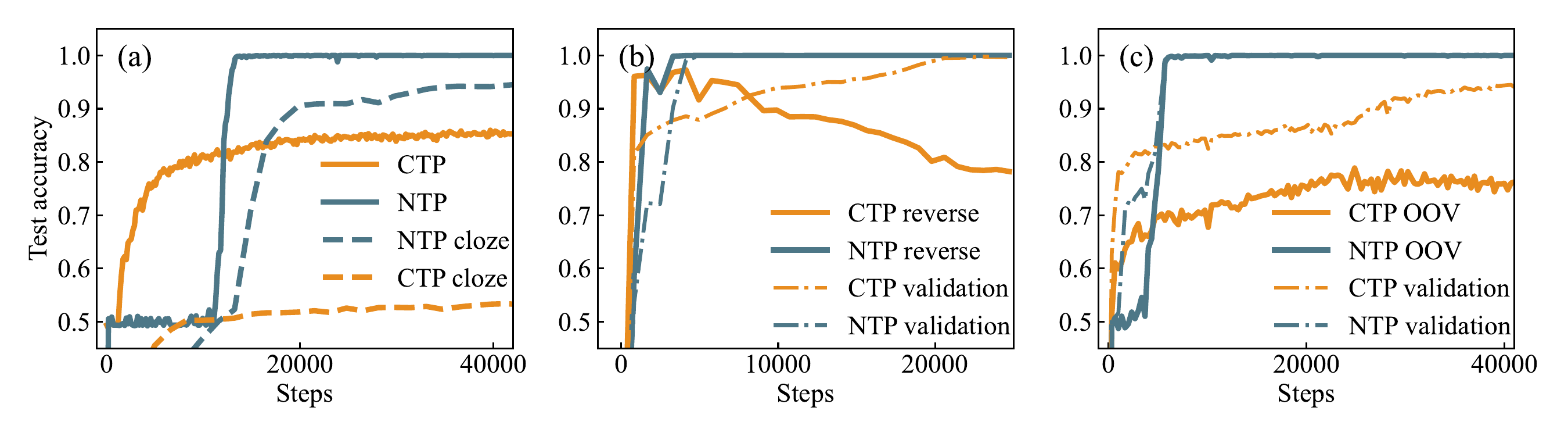}
  \caption{
(a) Accuracy of NTP and CTP on the original/cloze PrOntoQA task over training epochs. In the original task, NTP eventually achieves perfect accuracy, while CTP plateaus around 80\%. In the cloze task, the performance difference between NTP and CTP is enlarged. (b) 2-hop specific PrOntoQA: Performance of NTP and CTP on the specified key-answer PrOntoQA task. NTP maintains high accuracy without overfitting, whereas CTP overfits to the training data, leading to decreased accuracy on the reverse test set. (c) 1-hop specific PrOntoQA on OOV data: Accuracy of NTP and CTP on the 1-hop PrOntoQA task with OOV data. NTP achieves nearly 100\% accuracy, while CTP stabilizes around 70\%.
%(c)Performance of NTP and CTP on the PrOntoQA task using a pretrained GPT-2 model. CTP demonstrates faster learning speed compared to NTP.
}
    \label{fig:prontoqa_main}
\end{figure*}

Prior to delving into an in-depth analysis of NTP, it is imperative to establish a well-defined scope regarding which specific models and tasks are within the purview of this paper.
When discussing the reasoning capabilities of LLMs, natural language reasoning tasks should be taken into consideration, which encompasses natural language inference, multi-hop question answering, and commonsense reasoning \citep{survey_of_NLR}. As shown in Fig.~\ref{fig:add_flowchart}, these tasks are typically structured in a question-answer format, making them particularly well-suited for training using CTP, where the loss is calculated only on the answer tokens. Alternatively, one can employ NTP and compute the loss for the entire sentence. To ensure a fair comparison, we train the model from scratch (as opposed to using pretrained models with NTP). Consequently, commonsense reasoning tasks are excluded from the analysis. 

% Unless explicitly stated otherwise, the standard GPT-2 model \citep{radford2019language} with 12 layers and 12 heads is trained with reasoning tasks. More model configuration and training details can be found in the Appendix \ref{sec:experiment_detail}. 

\subsection{PrOntoQA task}

Inspired by ProofWriter \citep{tafjord-etal-2021-proofwriter}, PrOntoQA is a synthetic multi-hop inference dataset designed with simplistic grammar and unique proof paths. Each sequence consists of a set of facts and a question, requiring the language model to answer with either `True' or `False'. The random guess accuracy for this task is 50\%. We conducted comprehensive evaluations on the original PrOntoQA, while additionally proposing two modified datasets derived from this task, which are specifically designed to better support research on model generalization capabilities.

\paragraph{Original PrOntoQA task}

Following the experiment established in the \citep{PrOntoQA}, we implemented both NTP and CTP on the original PrOntoQA dataset.
Both training methods (NTP and CTP) easily surpass the random guess accuracy of 50\%. CTP initially learns the mappings effectively but stagnates at around 80\% accuracy. In contrast, NTP learns more slowly due to the presence of numerous noise terms but ultimately achieves 100\% accuracy, exhibiting an accuracy grokking phenomenon as shown in Fig.~\ref{fig:prontoqa_main} (a). 

% \begin{figure}[ht]
%     \centering
%     \includegraphics[width=0.7\columnwidth]{pics/0919_draft/Real_dataset/2hop_NO_FIX.pdf}
%     \caption{Accuracy of NTP and CTP on the original PrOntoQA task over training epochs. NTP eventually achieves perfect accuracy, while CTP plateaus around 80\%.}
%     \label{fig:2hop_no_fix}
% \end{figure}

\begin{figure}[!ht]
    \centering
    \includegraphics[width=0.6\columnwidth]{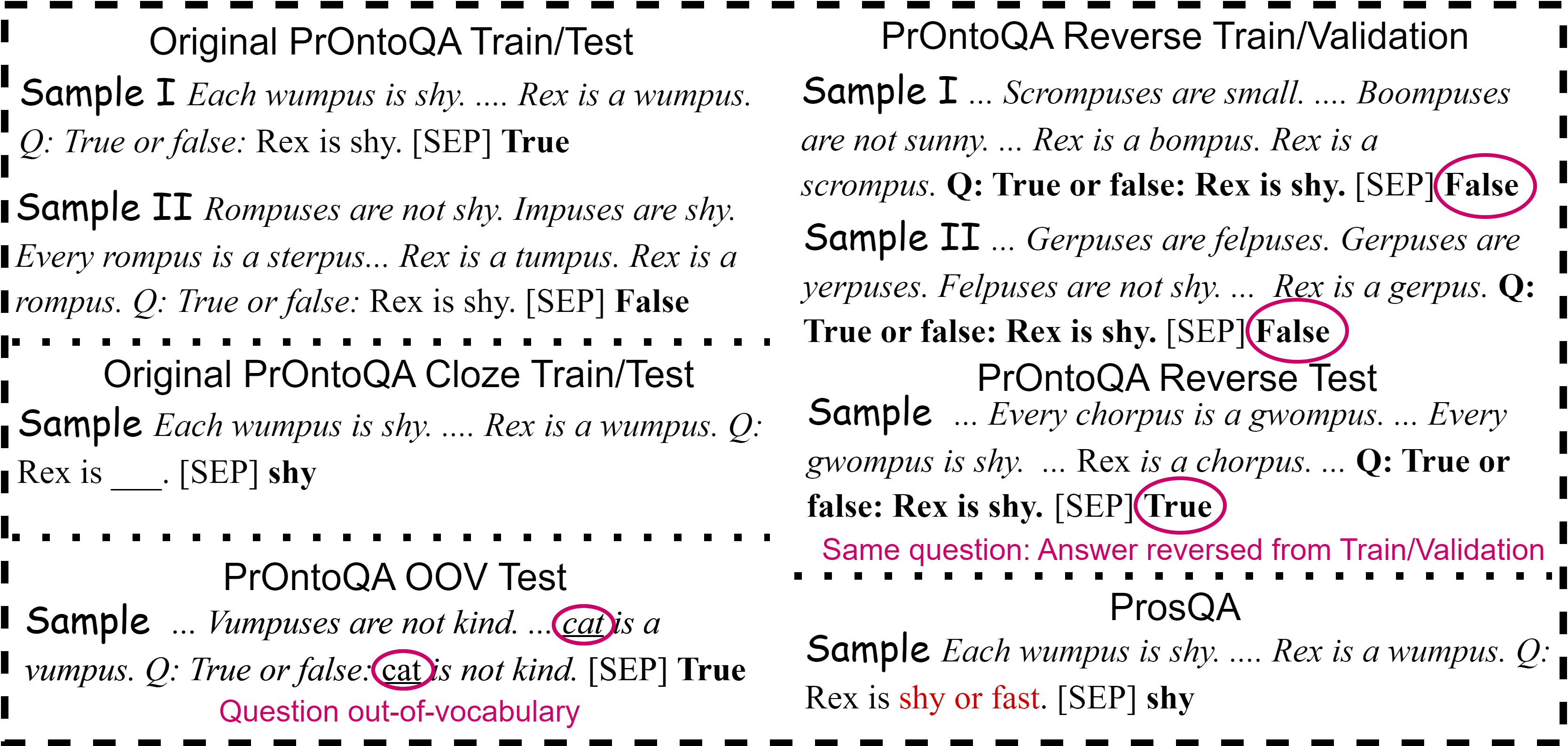}
    \caption{The description of different PrOntoQA tasks: Original, cloze, reverse, OOV test and its variation ProsQA \citep{coconut}. }
    \label{fig:prontoqa_flowchart}
\end{figure}

\paragraph{Reverse PrOntoQA task}

In the original PrOntoQA dataset, it is possible for two sequences to share the same question but have different answers due to random generation. This design choice allows the network to learn a common paradigm for solving reasoning tasks. To better assess the robustness differences between NTP and CTP, we made the following adjustments to the PrOntoQA dataset: Generated a series of samples with contradictory question-answer pairs, such as `Q: Lily is shy, A: True' and `Q: Lily is shy, A: False.' Then separate them into training/validation sets and reverse set. The generated sequences adhere to the reasoning principles. The detailed adjustment can be figured out in Fig.~\ref{fig:prontoqa_flowchart}.
% ii) Divide the sequences into Set1 and Set2. We ensure strict intra-dataset consistency while deliberately introducing inter-dataset contradictions across Set1 and Set2. iii) The training and validation sets were sampled from Set1, while the reverse set was exclusively drawn from Set2.

On the reverse set, we observe that while CTP enables the network to achieve an accuracy close to 1.0 initially, it rapidly overfits and begins to memorize the question-answer pairs from the training set, gradually forgetting the underlying reasoning rules. In contrast, NTP maintains an accuracy close to 1.0 over an extended period, demonstrating strong resistance to overfitting, as illustrated in Fig.~\ref{fig:prontoqa_main}(b).

\paragraph{OOV PrOntoQA task}
Based on the original construction of PrOntoQA, we introduce an Out-Of-Vocabulary (OOV) dataset, whose queries are not present in the training set. We evaluated the performance differences between models trained using NTP and CTP. Practically, NTP and CTP struggled with 2-hop reasoning on OOV data, we downgraded the dataset to 1-hop reasoning and replicated the experiments. The results, shown in Fig.~\ref{fig:prontoqa_main} (c), indicate that CTP maintains an accuracy of approximately 70\%, while NTP achieves nearly 100\% accuracy on the OOV dataset. This suggests that CTP is influenced by surface patterns in the data, whereas NTP effectively captures the underlying reasoning rules.

\subsection{Other natural language reasoning tasks}

In this work, except for PrOntoQA, we have meticulously curated a collection of reasoning datasets and implemented necessary preprocessing procedures to ensure data quality and suitability: LogicInference \citep{ontanon2022logicinferencenew}, CLUTRR \citep{sinha-etal-2019-clutrr}, Ruletaker \citep{ruletaker}, RobustLR \citep{sanyal-etal-2022-robustlr}, SimpleLogic \citep{SimpleLogic}, PARARULE Plus \citep{PARARULEPlus}, ReCOGS \citep{wu-etal-2023-recogs}, StepGame \citep{stepGame2022shi} and LogicAsker \citep{wan-etal-2024-logicasker}. Additionally, text classification tasks, including Yelp \citep{Yelp} and DBpedia \citep{dbpedia}, as well as the SNLI dataset \citep{SNLI}, are included in the comparison. The details of all these tasks could refer to Appendix \ref{sec:data_overview}.

\begin{figure*}[!htb]
    \centering
    \includegraphics[width=1\textwidth]{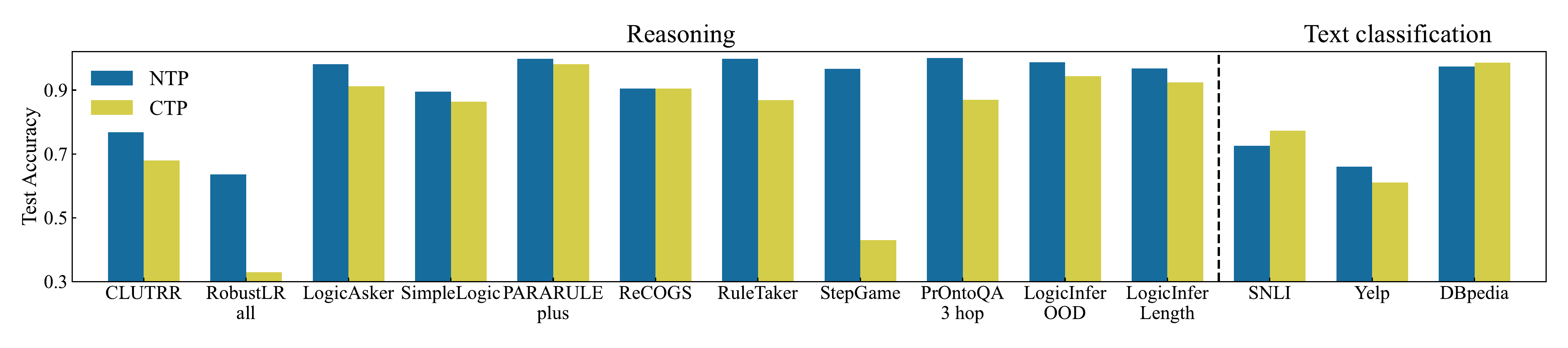}
    \caption{Performance comparison of NTP and CTP across various reasoning tasks. NTP consistently outperforms CTP in reasoning tasks, while performance on text classification tasks is more mixed. All the tasks are trained on the GPT-2 model (125M) from scratch to dismiss the effect of NTP in the pretraining stage. The accuracy is reported when the learning process becomes stable. }
    \label{fig:NLRtasks}
\end{figure*}

Our experimental results demonstrate that NTP exhibits superior performance compared to CTP across various reasoning-intensive tasks, including PrOntoQA, LogicAsker, and Ruletaker. Particularly noteworthy is NTP's exceptional capability in handling the challenging RobustLR task, where it partially captures underlying logical patterns, while CTP remains stagnant at random guess levels. As evidenced in Appendix \ref{sec:data_overview}, NTP demonstrates accelerated learning speed for tasks requiring strong reasoning capabilities. However, in text classification tasks that demand less sophisticated reasoning, the performance disparity between NTP and CTP diminishes significantly. In these scenarios, CTP exhibits a slight advantage in learning efficiency, as demonstrated by its comparable performance on SNLI and marginally better convergence rate on the DBpedia dataset. These findings are systematically presented and analyzed in Fig.~\ref{fig:NLRtasks}, which provides a comprehensive comparison of both approaches across different task categories.

\subsection{Anchor function: reasoning bias from complexity perspective}
Anchor function \citep{anchorfunction} is a novel synthetic dataset that distinguishes between the training data, test ID data, and OOD data. It provides a clear examination of the model's compositional generalization ability, training on composite operators, and trying to learn the atom operators. These operators are referred to as `anchors' while the target of anchors is notated as `keys'. 

We leverage four anchors $A,B,C,D$ and consider the 16 combinations of different anchor pairs, such as $AA, AB, \ldots, DD$. The specific operation of the single anchor is shifting. We set anchors $A, B, C, D$ shift the key $x\in[20,100]$ to $x+5, x+1, x-2, x-8$. The composite functions, like $AB$, means shifting $A$ then shifting $B$, so $AB(x) = B(A(x)) = x + 6$. After that, we pad the anchor-key pairs with uniform noise sequences, and the more detailed information could refer to Appendix \ref{sec:anchorfunction_appendix}.

\begin{figure}[ht]
\centering\includegraphics[width=0.6\columnwidth]{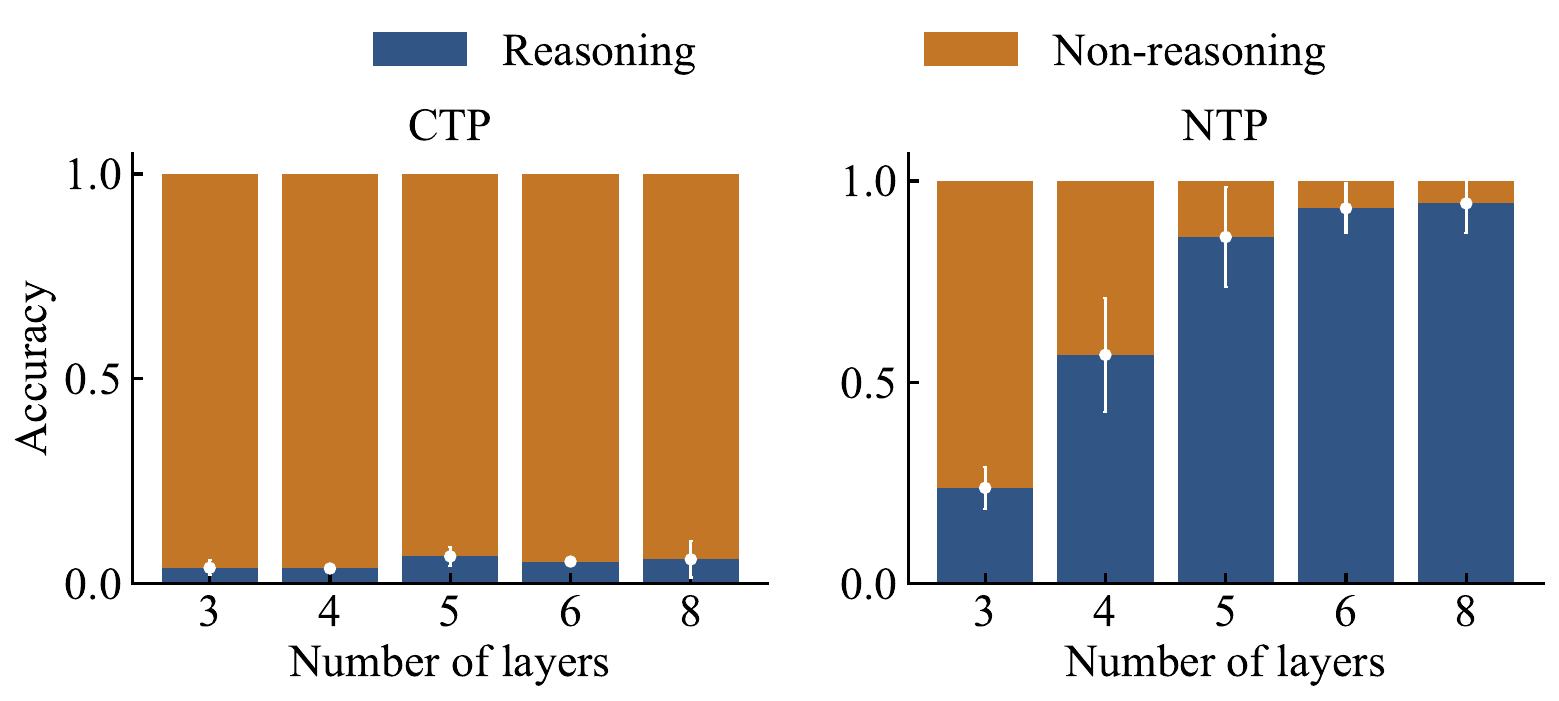}
    \caption{Accuracy on non-reasoning and reasoning solution of anchor function with different layers. The NTP could stably switch the non-reasoning solution to the reasoning solution. The
    error bars represent the standard deviation across 3-time runs on postnorm GPT2.}
    % 调整：放 4 个子图且 2468 只选择一列来绘制,vmin 到 vmax 为 0-1
    \label{fig:NTP Training process}
\end{figure}

An interesting question is, whether the transformer could learn the elementary functions only with composite functions. Here we set the model could see all composite pairs except $DC$, and convert the $CD$ into disturbance term $CD(x) := x - 6$. Two acceptable solutions exist in dataset. The reasoning solution is the model could learn elementary functions while treating pair $CD$ as an exception. If the model is biased by $CD$, leading to incorrect elementary functions, we call it a non-reasoning solution. From an intuitive perspective, the reasoning solution exhibits lower complexity, suggesting a model with superior generalization capabilities, whereas the non-reasoning solution presents contrasting characteristics to the inference with higher complexity.

In the \citep{zhangzhongwang}, authors have figured out the complexity, controlled by the initial scale, will affect the preference of the transformer. The smaller scale contributes to a more generalized model. However, the authors only focus on CTP training. Here we establish that, with the kaiming normal scale in which transformer should select the non-reasoning solution, will be shifted to a reasoning solution, as Fig.~\ref{fig:NTP Training process} shown. The NTP-trained models prefer reasoning solutions, and this tendency becomes increasingly evident as the depth of the model increases.

\section{NTP enhances early transfer generalization}
When the available data for a specific task is insufficient for training a model from scratch, transfer learning typically serves as an effective solution by finetuning a pretrained model with existing knowledge. In this section, we conduct transfer learning experiments between models trained using NTP and CTP across diverse downstream tasks. Our investigation yielded two results: (1) Models trained with NTP demonstrate accelerated generalization during the early stages of finetuning, although both approaches ultimately converge to comparable accuracy levels; (2) NTP-trained models exhibit a higher propensity for catastrophic forgetting during the finetuning process.
\begin{figure*}[!h]
    \centering
        \centering
        \includegraphics[width=\textwidth]{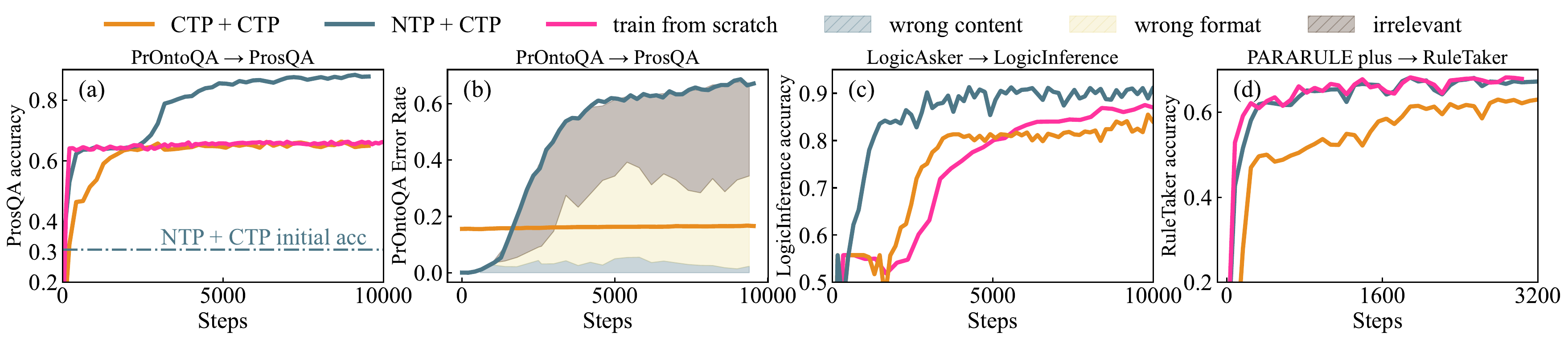}
    \caption{Finetuning results with multiple tasks. \textit{NTP+CTP} means the model is NTP-trained on previous task and CTP-finetuned on post task; \textit{CTP+CTP} means the model is CTP-trained on previous task and CTP-finetuned on post task. \textit{train from scratch} means the model is trained from scratch with the same configuration of CTP-funtuning.  (a, b) 2-hop PrOntoQA models continue to train on ProsQA. (a) The accuracy of ProsQA test data with the CTP finetuning process. (b) The accuracy of PrOntoQA test data and and the proportion of three error types of \textit{NTP+CTP} during finetuning. The \textit{wrong content}, \textit{wrong format} and \textit{irrelevant} represent incorrect answer content, improper answer formatting, and irrelevant responses. Regarding the omitted \textit{CTP+CTP} error types visualization, its \textit{wrong content} metric consistently maintains at 1.0, which demonstrates its immunity to finetuning perturbations. (c, d) More examples of transfer learning capability difference between NTP and CTP. }
    \label{fig:transfer2}
\end{figure*}

% \subsection{PrOntoQA to ProsQA}
% \begin{figure}[!h]
%     \centering
%         \centering
%         \includegraphics[width=0.7\columnwidth]{pics/0919_draft/Real_dataset/prontoqa2prosqa.pdf}
%     \caption{}
%     \label{fig:prontoqa2prosqa}
% \end{figure}
The ProsQA dataset, proposed in \citep{coconut}, represents an enhanced version of PrOntoQA, featuring more explicit reasoning graph structures. However, its limited scale precludes its use for training models from scratch. In this section, we primarily leverage its advantage of providing answer contrastive pairs to conduct finetuning experiments on models initially trained using both NTP and LTP on the 2-hop original PrOntoQA dataset. 

We employed a relatively low learning rate (2e-6) to meticulously capture the accuracy transitions between the original PrOntoQA 2-hop task and the new ProsQA task. The experimental results in Fig.~\ref{fig:transfer2} (a) demonstrate that the NTP model successfully predicts a portion of the validation set at the beginning, consistently outperforming CTP throughout the training process. This empirical evidence strongly suggests that NTP-trained models have inherent advantages for transfer learning applications.

However, in Fig.~\ref{fig:transfer2} (b), our empirical findings indicate that NTP-trained models are potentially more susceptible to catastrophic forgetting compared to their CTP counterparts. Through systematic evaluation, we observed a pronounced accuracy degradation on the original PrOntoQA dataset for NTP models as finetuning progressed, whereas CTP models showed only marginal performance decline, consistently maintaining a superior accuracy level.

Furthermore, We conducted an in-depth analysis of prediction errors, categorizing them into three distinct types: (1) Wrong content: instances where the model incorrectly predicts `False' when the ground truth is `True'; (2) Wrong format: cases such as responding with `shy' instead of the required `True/False' format to the question ``Is Rex shy?"; and (3) Irrelevant responses: The responses contains unrelated words from the input sentence. Our finding suggests the NTP-trained models are more willing to transfer the answer from PrOntoQA into new formats, ProsQA, while CTP-trained models demonstrate more consistent performance on PrOntoQA, even when the ProsQA task semantics remain identical. It treats the tasks separately and, as a consequence, shows weaker transfer ability. 

% \subsection{More transfer results}

Given the scale limitations of the dataset, we conducted additional experiments with multiple data groups to evaluate the transfer capabilities of NTP and CTP. Across various experimental settings, NTP consistently demonstrated superior transfer characteristics, even when the tasks were not directly related but shared similar reasoning patterns, as Fig.~\ref{fig:transfer2} (c, d) shows.

\section{Robustness of next token prediction}
In this section, we will investigate the robustness of NTP-trained models on the input level and the parameter level. Several studies on the robustness of transformer-based models have recently emerged. Compared with traditional language models or vision models, transformers are more robust in input noise in token level \citep{robustness_1, robustness_2, robustness_3, robustness_4}. However, with the complexity of real-world perturbation, transformer models exhibit significant room for improvement in terms of robustness \citep{robustness_reverse_1, robustness_reverse_2}. Especially, \citep{robustness_reverse_3} points out that the SFT procedure, which is similar to CTP, will do harm to the robustness of the NTP pretrained model. Meanwhile, we utilize the flatness as commonsense to explain the NTP generalization ability empirically. \citep{liu2023same} has shown that the better models' flatness, the better generalization ability on downstream tasks. 

\subsection{Embedding Noise}
\begin{figure*}[!h]
        \centering
        \includegraphics[width=\textwidth]{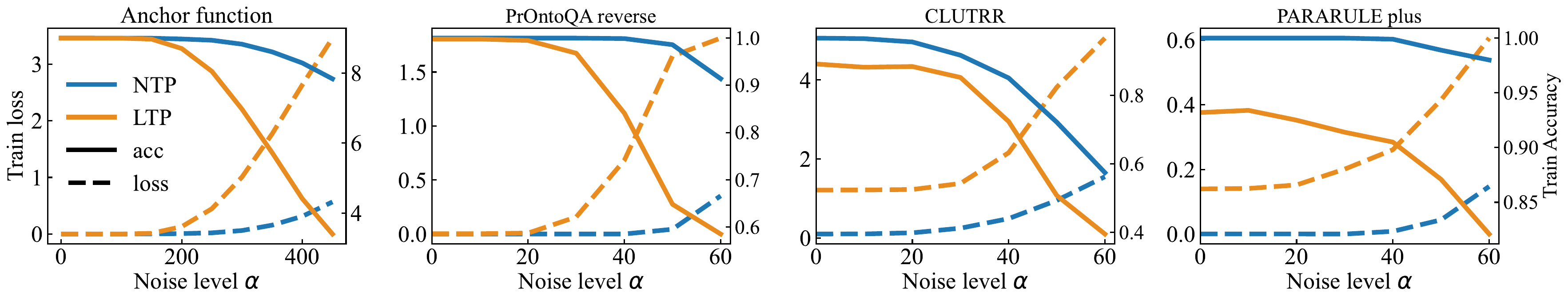}
    \caption{Effect of embedding noise on model performance in different reasoning tasks. The x-axis represents the perturbation strength $\alpha$ in Eq.~\eqref{eq:robustness}. NTP-trained models maintain higher accuracy under varying levels of input noise compared to CTP-trained models, which suffer from significant performance degradation in both accuracy and loss.}
    \label{fig:robustness}
\end{figure*}

The most straightforward approach to evaluating model robustness involves introducing controlled noise perturbations to the input data and quantitatively measuring the corresponding degradation in model accuracy. Following the settings applied in NETFune \citep{jain2023neftunenoisyembeddingsimprove}, noise restricted by the sequence length and model hidden size is added after the embedding layer as follows:
\begin{equation}
    \text{emb} \leftarrow \text{emb} + \frac{\alpha}{\sqrt{Sd}} \epsilon,
    \label{eq:robustness}
\end{equation}
where the noise $\epsilon$ is uniformly sampled from the range $[-1, 1]$, and $S, d$ represent for sequence length and embedding dimension separately. 

We have done a thorough analysis of the anchor function, as shown in Fig.~\ref{fig:robustness}(a), models trained with NTP are more stable under noise, while CTP-trained models exhibit high sensitivity. On the contrary of CTP, the NTP helps the model maintain its learned reasoning solution not only on the embedding layer, but on the output of different transformer blocks. With Fig.~\ref{fig:robustness}, on highly inference tasks like PrOntoQA and PARARULE plus, the performance patterns of NTP and CTP demonstrate remarkable similarity to their performance on anchor function.

\subsection{Train with misleading labels}
Another prevalent methodology for robustness evaluation involves deliberately introducing a proportion of noised samples into the training set, subsequently assessing the model's resilience to poisoned data. The addition is a tiny inference task widely used as a test set in the construction of new reasoning techniques of LLM \citep{deng2024explicitcotimplicitcot} and is the basic part of math reasoning steps \citep{ying2024internlmmathopenmathlarge}. 

Our addition dataset consists of addition problems bounded by 1000 and includes several random tokens corresponding to the random $x_i$ in the anchor function. When the length of the random token sequence is $n$, we denote it as Addition-R$n$. The numbers are padded to 4 digits and split into individual digits by the tokenizer. The total number of samples is $D = [0, 1000]^2$.

In the error addition task, we remove a square region from the center of $D$ with side length 100, denoted as $H = [400, 600]^2$. We randomly select 1000 or 2000 samples in $H$ and add noise $\pm 50$ to the labels, denoted them as poisoned samples $D_e$. The training set consists of $D \backslash H \cup D_e$, which includes the error samples, and the test dataset is $H \backslash D_e$. Drawing insights from our experience with anchor functions, we utilize an 8-layer transformer and observe the influence of poisoned samples.

\begin{figure}[!h]
    \centering
        \centering
        \includegraphics[width=0.7\columnwidth]{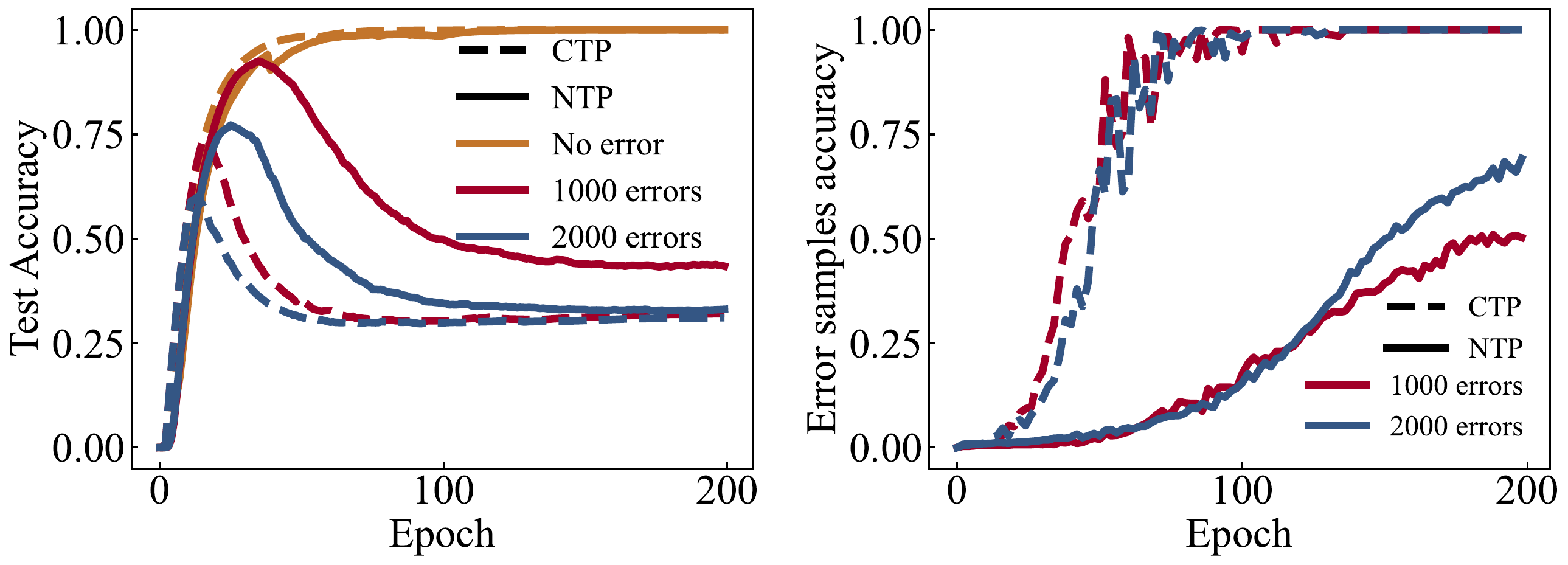}
    \caption{Comparison of NTP and CTP on the addition task with varying poisoned samples. The \textit{1000 errors} and \textit{2000 errors} denote training scenarios where an 800,000-sample dataset was deliberately contaminated with precisely 1,000 or 2,000 erroneous data points, respectively. (a) Test accuracy (on $H\backslash D_e$) (b) the memorizing speed of the poisoned samples $D_e$. The CTP could easily fit the errors before 100 epochs whereas NTP fits at a lower speed.}
    \label{fig:add task}
\end{figure}

Fig.~\ref{fig:add task} shows both NTP and CTP could easily learn addition without any poisoned samples. When the noise is introduced in the training data, NTP demonstrates superior performance, as evidenced by its higher peak test accuracy and delayed accuracy degradation compared to CTP. From Fig.~\ref{fig:add task} (b), meanwhile, CTP is trapped in memorizing poisoned samples at a faster speed than NTP. 

% 调整：robustness 放三个：anchor + add + clutrr 再加入一个 NTP 扰动后 CTP finetuing 一下,看看结果

\subsection{NTP reaches flatter minima}
\begin{figure*}[!ht]
        \centering
\includegraphics[width=0.75\textwidth]{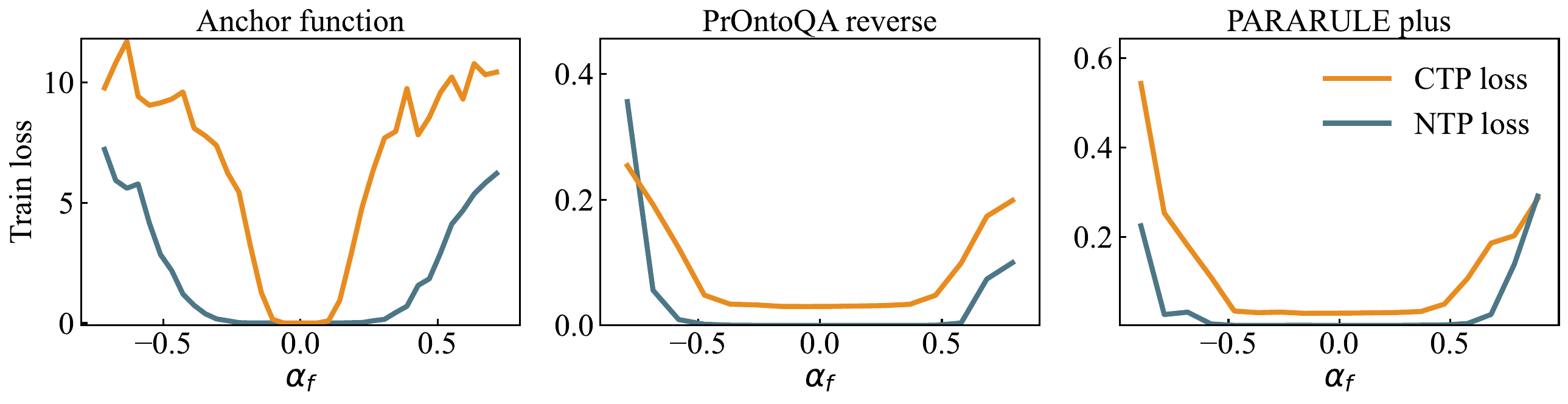}
    \caption{The flatness for task anchor function, PrOntoQA reverse, and PARARULE plus tasks. To alleviate the computational cost, the flatness is calculated on randomly sampled 20,000 instances from the training set. }
    \label{fig:flatness_main}
\end{figure*}

Flatness is commonsense to explain the NTP generalization ability starting with \citep{hochreiter1997flat} and is utilized to the neural network in \citep{keskar2017on}. 
The random direction approach \citep{visualize} is widely used to assess model robustness. The core involves selecting a direction from a normal distribution in the parameter space, followed by normalizing this direction according to the norm of the model parameters to ensure equitable comparison conditions. We denote the $\theta_{N}$ and $\theta_C$ as the NTP and CTP trained model, the $v$ as the random direction in parameter space, and the $\alpha_f$ as the intensity of perturbations. 
\begin{equation}
    \begin{aligned}
        \theta_N' &= \theta_N + \alpha_f \frac{\Vert \theta_N \Vert}{\Vert v \Vert} v, \\
        \theta_C' &= \theta_C + \alpha_f \frac{\Vert \theta_C \Vert}{\Vert v \Vert} v. \\
    \end{aligned}
\end{equation}

We tested the performance of NTP and CTP models under a moderate $\alpha_f$, which is shown in Fig.~\ref{fig:flatness_main}. For the simple anchor function task, the flatness of NTP overcomes CTP significantly. However, the flatness disparity between the two training approaches becomes considerably more nuanced in other tasks. The fact is the flatness analysis under this specific scenario favors CTP, primarily due to its constrained search space in reasoning tasks, where it only needs to discriminate between \textit{true} and \textit{false} responses. In contrast, NTP operates within a much larger search space, as it has to account for a diverse vocabulary during training.

\section{Discussion}
% 需要说明为什么在 SFT 阶段不应该使用 NTP 而应该使用 CTP, 两个原因：在 pretrain 过的模型上,已经继承了NTP原有的优秀平坦性等性质,因此不需要继续使用 NTP 来提高泛化能力；第二,SFT 阶段的指令微调等数据集存在单一的 instruction 的现象,实际上数据的表现很差,NTP 会造成反效果
\subsection{Relations between CTP and SFT}
\begin{figure}[!ht]
    \centering
    \includegraphics[width=0.6\columnwidth]{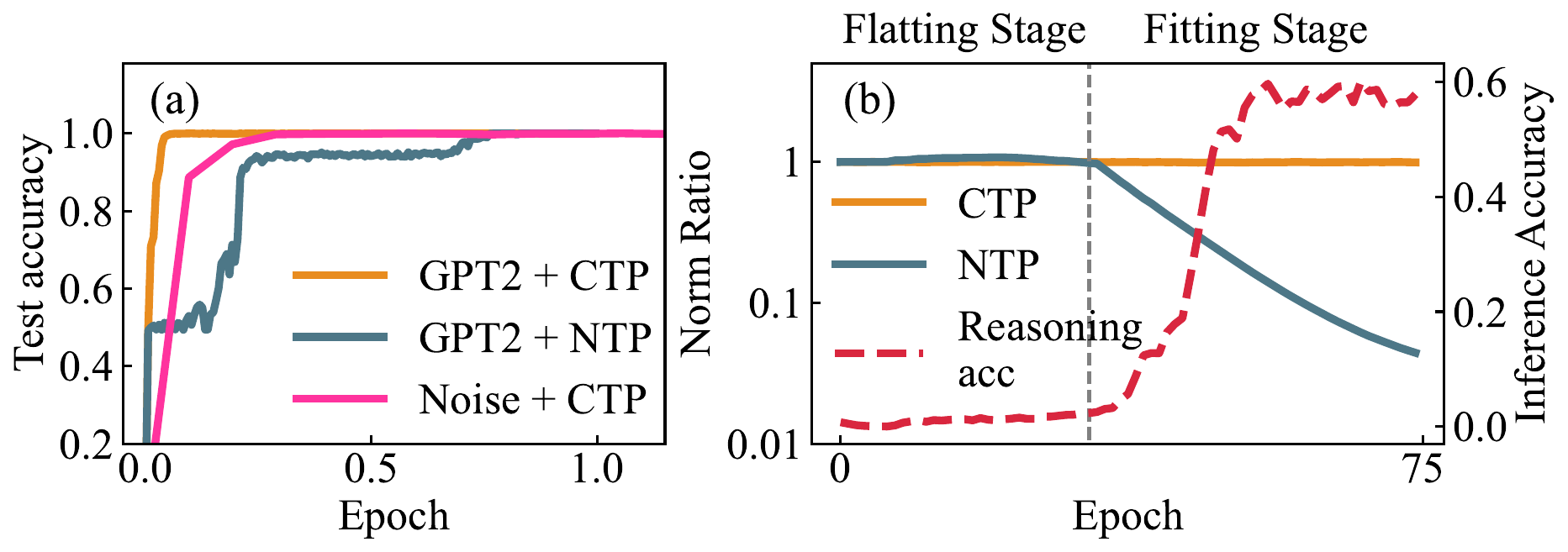}
    \caption{(a) The original 2-hop PrOntoQA task trained on the pretrained models. The legend entry \textit{GPT2} denotes the pretrained GPT-2 model parameters, and the \textit{Noise} denotes the model is pretrained on the noise term in PrOntoQA by NTP. (b) The ratio of gradient norm on random token position $t=T-1$ and critical token position $t=T$ of output in Eq.~\eqref{eq:loss}. The flatting stage and fitting stage are annotated, which corresponds with the reasoning accuracy raise.}
    \label{fig:pretrained_2hop}
\end{figure}

When adapting pretrained models for the downstream tasks, CTP (or SFT) is typically preferred over NTP. We evaluated the performance of NTP and CTP on the PrOntoQA dataset using a pretrained GPT-2 model. The results, depicted in Fig.~\ref{fig:pretrained_2hop}(a), show that CTP significantly outperforms NTP in terms of learning speed. This can be attributed to two factors: first, the pretrained model initialized through NTP already resides in a region of the loss landscape that is more amenable to generalization; second, pretraining endows the model with a certain level of reasoning capability. Consequently, additional noise in the corpus is unnecessary for aiding generalization, and the absence of noise allows the network to learn the mapping relationships more rapidly.

To our knowledge, the NTP loss function incorporates a component from CTP. We can isolate the CTP portion within the NTP loss and refer to the remaining part as the ``noise loss". Subsequently, we experimented with pretraining on both the anchor function task and PrOntoQA using this noise loss, followed by continued training with the CTP loss. This approach demonstrated improved generalization capability compared to directly training with CTP, evidenced by the reasoning solution obtained in the test accuracy reaching 100\% on PrOntoQA.
% A simple way to analysis the role of noise is to separate the training of $\mathcal{L}_{noise}$ and $\mathcal{L}_{C}$. We conducted several experiments on anchor function and PrOntoQA task. We optimize the noise loss  $\mathcal{L}_{noise}$ in pretraining stage, and CTP loss in post training stage. It shows the model gained better generalization capability than directly CTP, as a evidence of reasoning solution obtained in anchor function and test accuracy reaching 100\% in PrOntoQA. 

\subsection{Why models are not misled by noise}

% In this section, we aim to investigate the underlying factors contributing to NTP's enhanced robustness and superior generalization capabilities compared to CTP. Starting with the cleanest anchor function, the calculation loss on the noise term is the main difference. We decompose the NTP loss $\mathcal{L}_N$ into noise loss $\mathcal{L}_{noise}$ and CTP loss $\mathcal{L}_{C}$. 
Inspired by research on the performance of the BERT pretrained model with noisy data \citep{tänzer2022memorisationversusgeneralisationpretrained}, we noticed some phenomena associated with why NTP-trained models are not misled by noise terms. Using the clean anchor function as example, we discover that the gradient norm of the noise terms significantly decreased compared to the critical token (in Fig.~\ref{fig:pretrained_2hop} (b)), indicating that the network temporarily shifts its focus away from the noise during the learning process. The NTP learning process could be decomposed into two distinguishable stages: The flatting stage, the transformer trying to learn the distribution of the whole sequence. Fitting stage, after the $\mathcal{L}_{noise}$ reaches the lower bound entropy loss of NTP, the transformer notices the regularity of `key' item and gradually it turns to reasoning solution. 

\section*{Impact Statement}
In this study we proposed the reasoning bias phenomenon of next token prediction in the transformer-based models, after introducing critical token prediction approach in Q\&A datasets. Our research has clarified that even seemingly insignificant noise within sentences can serve as an effective regularizer in NTP training, simultaneously accelerating the model's reasoning capability acquisition. We firmly believe that this discovery poses no potential harm to human society while providing valuable insights into the comparative advantages of NTP over CTP. Furthermore, we underscore the necessity for deeper mechanistic analyses of NTP and exploration of alternative large model training paradigms, with special emphasis on investigating SFT's impact on model behavior.

\bibliography{bibs/ImplicitRegularization,bibs/next_token_prediction,bibs/references,bibs/intro}

\begin{thebibliography}{67}
\providecommand{\natexlab}[1]{#1}
\providecommand{\url}[1]{\texttt{#1}}
\expandafter\ifx\csname urlstyle\endcsname\relax
  \providecommand{\doi}[1]{doi: #1}\else
  \providecommand{\doi}{doi: \begingroup \urlstyle{rm}\Url}\fi

\bibitem[Abdin et~al.(2024)Abdin, Aneja, Behl, Bubeck, Eldan, Gunasekar, Harrison, Hewett, Javaheripi, Kauffmann, Lee, Lee, Li, Liu, Mendes, Nguyen, Price, de~Rosa, Saarikivi, Salim, Shah, Wang, Ward, Wu, Yu, Zhang, and Zhang]{abdin2024phi4technicalreport}
Abdin, M., Aneja, J., Behl, H., Bubeck, S., Eldan, R., Gunasekar, S., Harrison, M., Hewett, R.~J., Javaheripi, M., Kauffmann, P., Lee, J.~R., Lee, Y.~T., Li, Y., Liu, W., Mendes, C. C.~T., Nguyen, A., Price, E., de~Rosa, G., Saarikivi, O., Salim, A., Shah, S., Wang, X., Ward, R., Wu, Y., Yu, D., Zhang, C., and Zhang, Y.
\newblock Phi-4 technical report, 2024.
\newblock URL \url{https://arxiv.org/abs/2412.08905}.

\bibitem[Bachmann \& Nagarajan(2024)Bachmann and Nagarajan]{pmlr-v235-bachmann24a}
Bachmann, G. and Nagarajan, V.
\newblock The pitfalls of next-token prediction.
\newblock In Salakhutdinov, R., Kolter, Z., Heller, K., Weller, A., Oliver, N., Scarlett, J., and Berkenkamp, F. (eds.), \emph{Proceedings of the 41st International Conference on Machine Learning}, volume 235 of \emph{Proceedings of Machine Learning Research}, pp.\  2296--2318. PMLR, 21--27 Jul 2024.
\newblock URL \url{https://proceedings.mlr.press/v235/bachmann24a.html}.

\bibitem[Bank \& Giryes(2020)Bank and Giryes]{bank2020etf}
Bank, D. and Giryes, R.
\newblock An etf view of dropout regularization.
\newblock \emph{British Machine Vision Conference}, 2020.

\bibitem[Bao et~al.(2024)Bao, Peng, Hartill, Tan, Deng, Witbrock, and Liu]{PARARULEPlus}
Bao, Q., Peng, A.~Y., Hartill, T., Tan, N., Deng, Z., Witbrock, M., and Liu, J.
\newblock Multi-step deductive reasoning over natural language: An empirical study on out-of-distribution generalisation, 2024.
\newblock URL \url{https://arxiv.org/abs/2207.14000}.

\bibitem[Bhojanapalli et~al.(2021)Bhojanapalli, Chakrabarti, Glasner, Li, Unterthiner, and Veit]{robustness_3}
Bhojanapalli, S., Chakrabarti, A., Glasner, D., Li, D., Unterthiner, T., and Veit, A.
\newblock Understanding robustness of transformers for image classification.
\newblock In \emph{Proceedings of the IEEE/CVF International Conference on Computer Vision (ICCV)}, pp.\  10231--10241, October 2021.

\bibitem[Bowman et~al.(2015)Bowman, Angeli, Potts, and Manning]{SNLI}
Bowman, S.~R., Angeli, G., Potts, C., and Manning, C.~D.
\newblock A large annotated corpus for learning natural language inference.
\newblock In M{\`a}rquez, L., Callison-Burch, C., and Su, J. (eds.), \emph{Proceedings of the 2015 Conference on Empirical Methods in Natural Language Processing}, pp.\  632--642, Lisbon, Portugal, September 2015. Association for Computational Linguistics.
\newblock \doi{10.18653/v1/D15-1075}.
\newblock URL \url{https://aclanthology.org/D15-1075}.

\bibitem[Brown et~al.(2020)Brown, Mann, Ryder, Subbiah, Kaplan, Dhariwal, Neelakantan, Shyam, Sastry, Askell, Agarwal, Herbert-Voss, Krueger, Henighan, Child, Ramesh, Ziegler, Wu, Winter, Hesse, Chen, Sigler, teusz Litwin, Gray, Chess, Clark, Berner, McCandlish, Radford, Sutskever, and Amodei]{GPT3}
Brown, T.~B., Mann, B., Ryder, N., Subbiah, M., Kaplan, J., Dhariwal, P., Neelakantan, A., Shyam, P., Sastry, G., Askell, A., Agarwal, S., Herbert-Voss, A., Krueger, G., Henighan, T., Child, R., Ramesh, A., Ziegler, D.~M., Wu, J., Winter, C., Hesse, C., Chen, M., Sigler, E., teusz Litwin, M., Gray, S., Chess, B., Clark, J., Berner, C., McCandlish, S., Radford, A., Sutskever, I., and Amodei, D.
\newblock Language models are few-shot learners.
\newblock \emph{ArXiv}, abs/2005.14165, 2020.
\newblock URL \url{https://api.semanticscholar.org/CorpusID:218971783}.

\bibitem[Cavazza et~al.(2018)Cavazza, Morerio, Haeffele, Lane, Murino, and Vidal]{cavazza2018dropout}
Cavazza, J., Morerio, P., Haeffele, B., Lane, C., Murino, V., and Vidal, R.
\newblock Dropout as a low-rank regularizer for matrix factorization.
\newblock In \emph{International Conference on Artificial Intelligence and Statistics}, pp.\  435--444. PMLR, 2018.

\bibitem[Clark et~al.(2020)Clark, Tafjord, and Richardson]{ruletaker}
Clark, P., Tafjord, O., and Richardson, K.
\newblock Transformers as soft reasoners over language.
\newblock In Bessiere, C. (ed.), \emph{Proceedings of the Twenty-Ninth International Joint Conference on Artificial Intelligence, {IJCAI-20}}, pp.\  3882--3890. International Joint Conferences on Artificial Intelligence Organization, 7 2020.
\newblock \doi{10.24963/ijcai.2020/537}.
\newblock URL \url{https://doi.org/10.24963/ijcai.2020/537}.
\newblock Main track.

\bibitem[Deng et~al.(2024)Deng, Choi, and Shieber]{deng2024explicitcotimplicitcot}
Deng, Y., Choi, Y., and Shieber, S.
\newblock From explicit cot to implicit cot: Learning to internalize cot step by step, 2024.
\newblock URL \url{https://arxiv.org/abs/2405.14838}.

\bibitem[Fan et~al.(2019)Fan, Grave, and Joulin]{demand_dropout}
Fan, A., Grave, E., and Joulin, A.
\newblock Reducing transformer depth on demand with structured dropout, 2019.
\newblock URL \url{https://arxiv.org/abs/1909.11556}.

\bibitem[Feng \& Tu(2021)Feng and Tu]{feng2021inverse}
Feng, Y. and Tu, Y.
\newblock The inverse variance--flatness relation in stochastic gradient descent is critical for finding flat minima.
\newblock \emph{Proceedings of the National Academy of Sciences}, 118\penalty0 (9), 2021.

\bibitem[Gloeckle et~al.(2024)Gloeckle, Idrissi, Roziere, Lopez-Paz, and Synnaeve]{gloeckle2024better}
Gloeckle, F., Idrissi, B.~Y., Roziere, B., Lopez-Paz, D., and Synnaeve, G.
\newblock Better \& faster large language models via multi-token prediction.
\newblock In \emph{Forty-first International Conference on Machine Learning}, 2024.
\newblock URL \url{https://openreview.net/forum?id=pEWAcejiU2}.

\bibitem[Hao et~al.(2024)Hao, Sukhbaatar, Su, Li, Hu, Weston, and Tian]{coconut}
Hao, S., Sukhbaatar, S., Su, D., Li, X., Hu, Z., Weston, J., and Tian, Y.
\newblock Training large language models to reason in a continuous latent space, 2024.
\newblock URL \url{https://arxiv.org/abs/2412.06769}.

\bibitem[He \& Su(2024)He and Su]{heLawNextTokenPrediction2024}
He, H. and Su, W.~J.
\newblock A {{Law}} of {{Next-Token Prediction}} in {{Large Language Models}}, 2024.
\newblock URL \url{https://arxiv.org/abs/2408.13442v1}.

\bibitem[He et~al.(2024)He, Sun, Shen, and Li]{he2024matterstransformersattentionneeded}
He, S., Sun, G., Shen, Z., and Li, A.
\newblock What matters in transformers? not all attention is needed, 2024.
\newblock URL \url{https://arxiv.org/abs/2406.15786}.

\bibitem[Hendrycks et~al.(2020)Hendrycks, Liu, Wallace, Dziedzic, Krishnan, and Song]{robustness_2}
Hendrycks, D., Liu, X., Wallace, E., Dziedzic, A., Krishnan, R., and Song, D.
\newblock Pretrained transformers improve out-of-distribution robustness.
\newblock In Jurafsky, D., Chai, J., Schluter, N., and Tetreault, J. (eds.), \emph{Proceedings of the 58th Annual Meeting of the Association for Computational Linguistics}, pp.\  2744--2751, Online, July 2020. Association for Computational Linguistics.
\newblock \doi{10.18653/v1/2020.acl-main.244}.
\newblock URL \url{https://aclanthology.org/2020.acl-main.244/}.

\bibitem[Hochreiter \& Schmidhuber(1997)Hochreiter and Schmidhuber]{hochreiter1997flat}
Hochreiter, S. and Schmidhuber, J.
\newblock Flat minima.
\newblock \emph{Neural computation}, 9\penalty0 (1):\penalty0 1--42, 1997.

\bibitem[Jain et~al.(2023)Jain, yeh Chiang, Wen, Kirchenbauer, Chu, Somepalli, Bartoldson, Kailkhura, Schwarzschild, Saha, Goldblum, Geiping, and Goldstein]{jain2023neftunenoisyembeddingsimprove}
Jain, N., yeh Chiang, P., Wen, Y., Kirchenbauer, J., Chu, H.-M., Somepalli, G., Bartoldson, B.~R., Kailkhura, B., Schwarzschild, A., Saha, A., Goldblum, M., Geiping, J., and Goldstein, T.
\newblock Neftune: Noisy embeddings improve instruction finetuning, 2023.
\newblock URL \url{https://arxiv.org/abs/2310.05914}.

\bibitem[Keskar et~al.(2017)Keskar, Mudigere, Nocedal, Smelyanskiy, and Tang]{keskar2017on}
Keskar, N.~S., Mudigere, D., Nocedal, J., Smelyanskiy, M., and Tang, P. T.~P.
\newblock On large-batch training for deep learning: Generalization gap and sharp minima.
\newblock In \emph{International Conference on Learning Representations}, 2017.
\newblock URL \url{https://openreview.net/forum?id=H1oyRlYgg}.

\bibitem[Lehmann et~al.(2015)Lehmann, Isele, Jakob, Jentzsch, Kontokostas, Mendes, Hellmann, Morsey, Van~Kleef, Auer, et~al.]{dbpedia}
Lehmann, J., Isele, R., Jakob, M., Jentzsch, A., Kontokostas, D., Mendes, P.~N., Hellmann, S., Morsey, M., Van~Kleef, P., Auer, S., et~al.
\newblock Dbpedia--a large-scale, multilingual knowledge base extracted from wikipedia.
\newblock \emph{Semantic web}, 6\penalty0 (2):\penalty0 167--195, 2015.

\bibitem[Lengerich et~al.(2022)Lengerich, Xing, and Caruana]{lengerich2022dropout}
Lengerich, B.~J., Xing, E., and Caruana, R.
\newblock Dropout as a regularizer of interaction effects.
\newblock In \emph{International Conference on Artificial Intelligence and Statistics}, pp.\  7550--7564. PMLR, 2022.

\bibitem[Li et~al.(2023)Li, Hu, Nie, Han, Jiang, Guo, and Liu]{li2023dropkey}
Li, B., Hu, Y., Nie, X., Han, C., Jiang, X., Guo, T., and Liu, L.
\newblock Dropkey, 2023.
\newblock URL \url{https://arxiv.org/abs/2208.02646}.

\bibitem[Li et~al.(2024{\natexlab{a}})Li, Tian, Zerong, Song, and Xia]{robustness_4}
Li, C., Tian, Y., Zerong, Z., Song, Y., and Xia, F.
\newblock Challenging large language models with new tasks: A study on their adaptability and robustness.
\newblock In Ku, L.-W., Martins, A., and Srikumar, V. (eds.), \emph{Findings of the Association for Computational Linguistics: ACL 2024}, pp.\  8140--8162, Bangkok, Thailand, August 2024{\natexlab{a}}. Association for Computational Linguistics.
\newblock \doi{10.18653/v1/2024.findings-acl.485}.
\newblock URL \url{https://aclanthology.org/2024.findings-acl.485/}.

\bibitem[Li et~al.(2018)Li, Xu, Taylor, Studer, and Goldstein]{visualize}
Li, H., Xu, Z., Taylor, G., Studer, C., and Goldstein, T.
\newblock Visualizing the loss landscape of neural nets.
\newblock In Bengio, S., Wallach, H., Larochelle, H., Grauman, K., Cesa-Bianchi, N., and Garnett, R. (eds.), \emph{Advances in Neural Information Processing Systems}, volume~31. Curran Associates, Inc., 2018.
\newblock URL \url{https://proceedings.neurips.cc/paper_files/paper/2018/file/a41b3bb3e6b050b6c9067c67f663b915-Paper.pdf}.

\bibitem[Li et~al.(2024{\natexlab{b}})Li, Huang, Ildiz, Singh~Rawat, and Oymak]{MechanicsofNextTokenPredictionwithSelf-Attention}
Li, Y., Huang, Y., Ildiz, M.~E., Singh~Rawat, A., and Oymak, S.
\newblock Mechanics of next token prediction with self-attention.
\newblock In Dasgupta, S., Mandt, S., and Li, Y. (eds.), \emph{Proceedings of The 27th International Conference on Artificial Intelligence and Statistics}, volume 238 of \emph{Proceedings of Machine Learning Research}, pp.\  685--693. PMLR, 02--04 May 2024{\natexlab{b}}.
\newblock URL \url{https://proceedings.mlr.press/v238/li24f.html}.

\bibitem[Lin et~al.(2025)Lin, Gou, Gong, Liu, Shen, Xu, Lin, Yang, Jiao, Duan, and Chen]{rho1}
Lin, Z., Gou, Z., Gong, Y., Liu, X., Shen, Y., Xu, R., Lin, C., Yang, Y., Jiao, J., Duan, N., and Chen, W.
\newblock Rho-1: Not all tokens are what you need, 2025.
\newblock URL \url{https://arxiv.org/abs/2404.07965}.

\bibitem[Liu et~al.(2023)Liu, Xie, Li, and Ma]{liu2023same}
Liu, H., Xie, S.~M., Li, Z., and Ma, T.
\newblock Same pre-training loss, better downstream: Implicit bias matters for language models.
\newblock In \emph{International Conference on Machine Learning}, pp.\  22188--22214. PMLR, 2023.

\bibitem[Ma \& Ying(2021)Ma and Ying]{ma2021linear}
Ma, C. and Ying, L.
\newblock On linear stability of sgd and input-smoothness of neural networks.
\newblock \emph{Advances in Neural Information Processing Systems}, 34:\penalty0 16805--16817, 2021.

\bibitem[Madden et~al.(2024)Madden, Fox, and Thrampoulidis]{madden2024nexttokenpredictioncapacitygeneral}
Madden, L., Fox, C., and Thrampoulidis, C.
\newblock Next-token prediction capacity: general upper bounds and a lower bound for transformers, 2024.
\newblock URL \url{https://arxiv.org/abs/2405.13718}.

\bibitem[Mianjy et~al.(2018)Mianjy, Arora, and Vidal]{mianjy2018implicit}
Mianjy, P., Arora, R., and Vidal, R.
\newblock On the implicit bias of dropout.
\newblock In \emph{International Conference on Machine Learning}, pp.\  3540--3548. PMLR, 2018.

\bibitem[Mishra et~al.(2022)Mishra, Sachdeva, and Baral]{robustness_reverse_2}
Mishra, S., Sachdeva, B.~S., and Baral, C.
\newblock Pretrained transformers do not always improve robustness, 2022.
\newblock URL \url{https://arxiv.org/abs/2210.07663}.

\bibitem[Moradi \& Samwald(2021)Moradi and Samwald]{robustness_reverse_1}
Moradi, M. and Samwald, M.
\newblock Evaluating the robustness of neural language models to input perturbations.
\newblock In Moens, M.-F., Huang, X., Specia, L., and Yih, S. W.-t. (eds.), \emph{Proceedings of the 2021 Conference on Empirical Methods in Natural Language Processing}, pp.\  1558--1570, Online and Punta Cana, Dominican Republic, November 2021. Association for Computational Linguistics.
\newblock \doi{10.18653/v1/2021.emnlp-main.117}.
\newblock URL \url{https://aclanthology.org/2021.emnlp-main.117/}.

\bibitem[Mori et~al.(2021)Mori, Ziyin, Liu, and Ueda]{mori2021power}
Mori, T., Ziyin, L., Liu, K., and Ueda, M.
\newblock Power-law escape rate of sgd.
\newblock \emph{arXiv preprint arXiv:2105.09557}, 2021.

\bibitem[Ontanon et~al.(2022)Ontanon, Ainslie, Cvicek, and Fisher]{ontanon2022logicinferencenew}
Ontanon, S., Ainslie, J., Cvicek, V., and Fisher, Z.
\newblock Logicinference: A new dataset for teaching logical inference to seq2seq models, 2022.
\newblock URL \url{https://arxiv.org/abs/2203.15099}.

\bibitem[Radford \& Narasimhan(2018)Radford and Narasimhan]{GPT1}
Radford, A. and Narasimhan, K.
\newblock Improving language understanding by generative pre-training.
\newblock 2018.
\newblock URL \url{https://api.semanticscholar.org/CorpusID:49313245}.

\bibitem[Radford et~al.(2019)Radford, Wu, Child, Luan, Amodei, and Sutskever]{GPT2}
Radford, A., Wu, J., Child, R., Luan, D., Amodei, D., and Sutskever, I.
\newblock Language models are unsupervised multitask learners.
\newblock 2019.

\bibitem[Sanyal et~al.(2022)Sanyal, Liao, and Ren]{sanyal-etal-2022-robustlr}
Sanyal, S., Liao, Z., and Ren, X.
\newblock {R}obust{LR}: A diagnostic benchmark for evaluating logical robustness of deductive reasoners.
\newblock In Goldberg, Y., Kozareva, Z., and Zhang, Y. (eds.), \emph{Proceedings of the 2022 Conference on Empirical Methods in Natural Language Processing}, pp.\  9614--9631, Abu Dhabi, United Arab Emirates, December 2022. Association for Computational Linguistics.
\newblock \doi{10.18653/v1/2022.emnlp-main.653}.
\newblock URL \url{https://aclanthology.org/2022.emnlp-main.653}.

\bibitem[Saparov \& He(2023)Saparov and He]{PrOntoQA}
Saparov, A. and He, H.
\newblock Language models are greedy reasoners: A systematic formal analysis of chain-of-thought.
\newblock In \emph{The Eleventh International Conference on Learning Representations}, 2023.
\newblock URL \url{https://openreview.net/forum?id=qFVVBzXxR2V}.

\bibitem[Shi et~al.(2022)Shi, Zhang, and Lipani]{stepGame2022shi}
Shi, Z., Zhang, Q., and Lipani, A.
\newblock Stepgame: A new benchmark for robust multi-hop spatial reasoning in texts.
\newblock In \emph{Proceedings of the AAAI Conference on Artificial Intelligence}, volume~36, pp.\  11321--11329, Jun. 2022.
\newblock \doi{10.1609/aaai.v36i10.21383}.
\newblock URL \url{https://ojs.aaai.org/index.php/AAAI/article/view/21383}.

\bibitem[Sinha et~al.(2019)Sinha, Sodhani, Dong, Pineau, and Hamilton]{sinha-etal-2019-clutrr}
Sinha, K., Sodhani, S., Dong, J., Pineau, J., and Hamilton, W.~L.
\newblock {CLUTRR}: A diagnostic benchmark for inductive reasoning from text.
\newblock In Inui, K., Jiang, J., Ng, V., and Wan, X. (eds.), \emph{Proceedings of the 2019 Conference on Empirical Methods in Natural Language Processing and the 9th International Joint Conference on Natural Language Processing (EMNLP-IJCNLP)}, pp.\  4506--4515, Hong Kong, China, November 2019. Association for Computational Linguistics.
\newblock \doi{10.18653/v1/D19-1458}.
\newblock URL \url{https://aclanthology.org/D19-1458}.

\bibitem[Tafjord et~al.(2021)Tafjord, Dalvi, and Clark]{tafjord-etal-2021-proofwriter}
Tafjord, O., Dalvi, B., and Clark, P.
\newblock {P}roof{W}riter: Generating implications, proofs, and abductive statements over natural language.
\newblock In Zong, C., Xia, F., Li, W., and Navigli, R. (eds.), \emph{Findings of the Association for Computational Linguistics: ACL-IJCNLP 2021}, pp.\  3621--3634, Online, August 2021. Association for Computational Linguistics.
\newblock \doi{10.18653/v1/2021.findings-acl.317}.
\newblock URL \url{https://aclanthology.org/2021.findings-acl.317}.

\bibitem[T{\"a}nzer et~al.(2022)T{\"a}nzer, Ruder, and Rei]{tänzer2022memorisationversusgeneralisationpretrained}
T{\"a}nzer, M., Ruder, S., and Rei, M.
\newblock Memorisation versus generalisation in pre-trained language models.
\newblock In Muresan, S., Nakov, P., and Villavicencio, A. (eds.), \emph{Proceedings of the 60th Annual Meeting of the Association for Computational Linguistics (Volume 1: Long Papers)}, pp.\  7564--7578, Dublin, Ireland, May 2022. Association for Computational Linguistics.
\newblock \doi{10.18653/v1/2022.acl-long.521}.
\newblock URL \url{https://aclanthology.org/2022.acl-long.521}.

\bibitem[Thrampoulidis(2024)]{thrampoulidisimplicit}
Thrampoulidis, C.
\newblock Implicit optimization bias of next-token prediction in linear models.
\newblock In \emph{The Thirty-eighth Annual Conference on Neural Information Processing Systems}, 2024.

\bibitem[Tu et~al.(2020)Tu, Lalwani, Gella, and He]{robustness_1}
Tu, L., Lalwani, G., Gella, S., and He, H.
\newblock An empirical study on robustness to spurious correlations using pre-trained language models.
\newblock \emph{Transactions of the Association for Computational Linguistics}, 8:\penalty0 621--633, 10 2020.
\newblock ISSN 2307-387X.
\newblock \doi{10.1162/tacl_a_00335}.
\newblock URL \url{https://doi.org/10.1162/tacl\_a\_00335}.

\bibitem[Wan et~al.(2024)Wan, Wang, Yang, Yuan, Huang, He, Jiao, and Lyu]{wan-etal-2024-logicasker}
Wan, Y., Wang, W., Yang, Y., Yuan, Y., Huang, J.-t., He, P., Jiao, W., and Lyu, M.
\newblock {L}ogic{A}sker: Evaluating and improving the logical reasoning ability of large language models.
\newblock In Al-Onaizan, Y., Bansal, M., and Chen, Y.-N. (eds.), \emph{Proceedings of the 2024 Conference on Empirical Methods in Natural Language Processing}, pp.\  2124--2155, Miami, Florida, USA, November 2024. Association for Computational Linguistics.
\newblock \doi{10.18653/v1/2024.emnlp-main.128}.
\newblock URL \url{https://aclanthology.org/2024.emnlp-main.128}.

\bibitem[Wang et~al.(2023)Wang, Ma, Yu, Gui, Zhang, Huang, Ma, Chang, Zhang, Shen, Wang, Zhao, and Tao]{robustness_reverse_3}
Wang, H., Ma, G., Yu, C., Gui, N., Zhang, L., Huang, Z., Ma, S., Chang, Y., Zhang, S., Shen, L., Wang, X., Zhao, P., and Tao, D.
\newblock Are large language models really robust to word-level perturbations?
\newblock In \emph{Socially Responsible Language Modelling Research}, 2023.
\newblock URL \url{https://openreview.net/forum?id=mVhOKo62Q2}.

\bibitem[Wei et~al.(2020)Wei, Kakade, and Ma]{wei2020implicit}
Wei, C., Kakade, S., and Ma, T.
\newblock The implicit and explicit regularization effects of dropout.
\newblock In \emph{International Conference on Machine Learning}, pp.\  10181--10192. PMLR, 2020.

\bibitem[Weston et~al.(2015)Weston, Bordes, Chopra, Rush, van Merriënboer, Joulin, and Mikolov]{babi}
Weston, J., Bordes, A., Chopra, S., Rush, A.~M., van Merriënboer, B., Joulin, A., and Mikolov, T.
\newblock Towards ai-complete question answering: A set of prerequisite toy tasks, 2015.
\newblock URL \url{https://arxiv.org/abs/1502.05698}.

\bibitem[Wu et~al.(2020)Wu, Hu, Xiong, Huan, Braverman, and Zhu]{wu2020noisy}
Wu, J., Hu, W., Xiong, H., Huan, J., Braverman, V., and Zhu, Z.
\newblock On the noisy gradient descent that generalizes as sgd.
\newblock In \emph{International Conference on Machine Learning}, pp.\  10367--10376. PMLR, 2020.

\bibitem[Wu et~al.(2018)Wu, Ma, et~al.]{wu2018sgd}
Wu, L., Ma, C., et~al.
\newblock How sgd selects the global minima in over-parameterized learning: A dynamical stability perspective.
\newblock \emph{Advances in Neural Information Processing Systems}, 31, 2018.

\bibitem[Wu et~al.(2021)Wu, Wu, Meng, Xia, Xie, Qin, Dai, and Liu]{UniDrop}
Wu, Z., Wu, L., Meng, Q., Xia, Y., Xie, S., Qin, T., Dai, X., and Liu, T.-Y.
\newblock Unidrop: A simple yet effective technique to improve transformer without extra cost, 2021.
\newblock URL \url{https://arxiv.org/abs/2104.04946}.

\bibitem[Wu et~al.(2023)Wu, Manning, and Potts]{wu-etal-2023-recogs}
Wu, Z., Manning, C.~D., and Potts, C.
\newblock {R}e{COGS}: How incidental details of a logical form overshadow an evaluation of semantic interpretation.
\newblock \emph{Transactions of the Association for Computational Linguistics}, 11:\penalty0 1719--1733, 2023.
\newblock \doi{10.1162/tacl_a_00623}.
\newblock URL \url{https://aclanthology.org/2023.tacl-1.96}.

\bibitem[Xie et~al.(2020)Xie, Sato, and Sugiyama]{xie2020diffusion}
Xie, Z., Sato, I., and Sugiyama, M.
\newblock A diffusion theory for deep learning dynamics: Stochastic gradient descent exponentially favors flat minima.
\newblock \emph{arXiv preprint arXiv:2002.03495}, 2020.

\bibitem[Yelp Dataset()]{Yelp}
Yelp Dataset, 2014.
\newblock URL \url{http://www.yelp.com/ dataset_challenge}.

\bibitem[Ying et~al.(2024)Ying, Zhang, Li, Zhou, Shao, Fei, Ma, Hong, Liu, Wang, Wang, Wu, Li, Zhou, Liu, Zhang, Zhang, Yan, Qiu, Wang, Chen, and Lin]{ying2024internlmmathopenmathlarge}
Ying, H., Zhang, S., Li, L., Zhou, Z., Shao, Y., Fei, Z., Ma, Y., Hong, J., Liu, K., Wang, Z., Wang, Y., Wu, Z., Li, S., Zhou, F., Liu, H., Zhang, S., Zhang, W., Yan, H., Qiu, X., Wang, J., Chen, K., and Lin, D.
\newblock Internlm-math: Open math large language models toward verifiable reasoning, 2024.
\newblock URL \url{https://arxiv.org/abs/2402.06332}.

\bibitem[Yu et~al.(2024)Yu, Zhang, Tiwari, and Wang]{survey_of_NLR}
Yu, F., Zhang, H., Tiwari, P., and Wang, B.
\newblock Natural language reasoning, a survey.
\newblock \emph{ACM Comput. Surv.}, 56\penalty0 (12), October 2024.
\newblock ISSN 0360-0300.
\newblock \doi{10.1145/3664194}.
\newblock URL \url{https://doi.org/10.1145/3664194}.

\bibitem[Zehui et~al.(2019)Zehui, Liu, Huang, Chen, Qiu, and Huang]{dropattention}
Zehui, L., Liu, P., Huang, L., Chen, J., Qiu, X., and Huang, X.
\newblock Dropattention: A regularization method for fully-connected self-attention networks, 2019.
\newblock URL \url{https://arxiv.org/abs/1907.11065}.

\bibitem[Zhang et~al.(2022)Zhang, Qu, Shao, and Yang]{dropdim}
Zhang, H., Qu, D., Shao, K., and Yang, X.
\newblock Dropdim: A regularization method for transformer networks.
\newblock \emph{IEEE Signal Processing Letters}, 29:\penalty0 474--478, 2022.
\newblock \doi{10.1109/LSP.2022.3140693}.

\bibitem[Zhang et~al.(2023{\natexlab{a}})Zhang, Li, Meng, Chang, and Van~den Broeck]{SimpleLogic}
Zhang, H., Li, L.~H., Meng, T., Chang, K.-W., and Van~den Broeck, G.
\newblock On the paradox of learning to reason from data.
\newblock In Elkind, E. (ed.), \emph{Proceedings of the Thirty-Second International Joint Conference on Artificial Intelligence, {IJCAI-23}}, pp.\  3365--3373. International Joint Conferences on Artificial Intelligence Organization, 8 2023{\natexlab{a}}.
\newblock \doi{10.24963/ijcai.2023/375}.
\newblock URL \url{https://doi.org/10.24963/ijcai.2023/375}.
\newblock Main Track.

\bibitem[Zhang \& Xu(2024)Zhang and Xu]{zhang2024implicit}
Zhang, Z. and Xu, Z.-Q.~J.
\newblock Implicit regularization of dropout.
\newblock \emph{IEEE Transactions on Pattern Analysis and Machine Intelligence}, 2024.

\bibitem[Zhang et~al.(2023{\natexlab{b}})Zhang, Li, Luo, and Xu]{zhang2023stochastic}
Zhang, Z., Li, Y., Luo, T., and Xu, Z.-Q.~J.
\newblock Stochastic modified equations and dynamics of dropout algorithm.
\newblock \emph{arXiv preprint arXiv:2305.15850}, 2023{\natexlab{b}}.

\bibitem[Zhang et~al.(2024{\natexlab{a}})Zhang, Lin, Wang, Zhang, and Xu]{zhangzhongwang}
Zhang, Z., Lin, P., Wang, Z., Zhang, Y., and Xu, Z.-Q.~J.
\newblock Initialization is critical to whether transformers fit composite functions by inference or memorizing, 2024{\natexlab{a}}.
\newblock URL \url{https://arxiv.org/abs/2405.05409}.

\bibitem[Zhang et~al.(2024{\natexlab{b}})Zhang, Wang, Yao, Zhou, Li, E, and Xu]{anchorfunction}
Zhang, Z., Wang, Z., Yao, J., Zhou, Z., Li, X., E, W., and Xu, Z.-Q.~J.
\newblock Anchor function: a type of benchmark functions for studying language models, 2024{\natexlab{b}}.
\newblock URL \url{https://arxiv.org/abs/2401.08309}.

\bibitem[Zhao et~al.(2024)Zhao, Behnia, Vakilian, and Thrampoulidis]{zhao2024implicit}
Zhao, Y., Behnia, T., Vakilian, V., and Thrampoulidis, C.
\newblock Implicit geometry of next-token prediction: From language sparsity patterns to model representations.
\newblock In \emph{First Conference on Language Modeling}, 2024.
\newblock URL \url{https://openreview.net/forum?id=qyilOnIRHI}.

\bibitem[Zhou et~al.(2020)Zhou, Ge, Wei, Zhou, and Xu]{drophead}
Zhou, W., Ge, T., Wei, F., Zhou, M., and Xu, K.
\newblock Scheduled {D}rop{H}ead: A regularization method for transformer models.
\newblock In Cohn, T., He, Y., and Liu, Y. (eds.), \emph{Findings of the Association for Computational Linguistics: EMNLP 2020}, pp.\  1971--1980, Online, November 2020. Association for Computational Linguistics.
\newblock \doi{10.18653/v1/2020.findings-emnlp.178}.
\newblock URL \url{https://aclanthology.org/2020.findings-emnlp.178}.

\bibitem[Zhu et~al.(2019)Zhu, Wu, Yu, Wu, and Ma]{zhu2019anisotropic}
Zhu, Z., Wu, J., Yu, B., Wu, L., and Ma, J.
\newblock The anisotropic noise in stochastic gradient descent: Its behavior of escaping from sharp minima and regularization effects.
\newblock In \emph{International Conference on Machine Learning}, pp.\  7654--7663. PMLR, 2019.

\end{thebibliography}
\bibliographystyle{icml2025}
\newpage
\section*{Appendix}
% \appendix

\subsection{The reasoning task datasets overview}
\label{sec:data_overview}
\begin{figure}[!ht]
    \centering
    \includegraphics[width=0.8\linewidth]{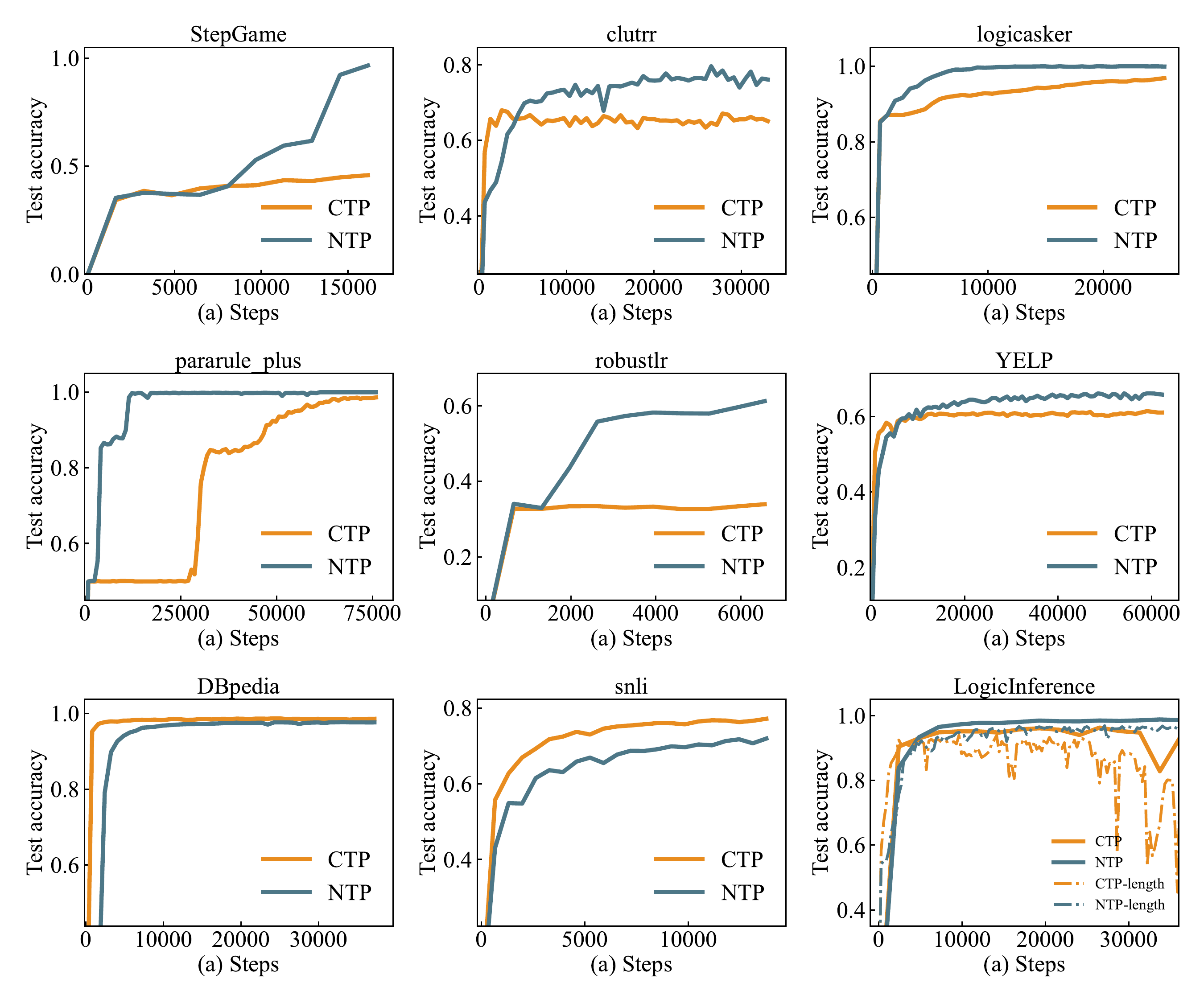}
    \caption{The NTP and CTP training process of reasoning tasks and text classification tasks. CTP outperforms NTP on tasks that involve shorter texts and require less extensive reasoning like DBpedia or SNLI but NTP outperforms CTP on reasoning data, such as PrOntoQA, RobustLR etc. The first figure PrOntoQA-2hop cloze means the accuracy of cloze version about PrOntoQA.}
    \label{fig:allNLR}
\end{figure}
\paragraph{PrOntoQA \& ProsQA}
Every sequence in PrOntoQA dataset consists of three parts: fact, question and answer. Some noise disturbance terms are mixed in the fact part. An example of 1-hop reasoning is below:
\begin{displayquote}
        Fact: \\Every gwompus is not amenable. Every gwompus is a chorpus. Gwompuses are zhorpuses. Every chorpus is transparent. Chorpuses are gerpuses. Every chorpus is a storpus. Gerpuses are not hot. Gerpuses are bompuses. Each gerpus is a boompus. Bompuses are sweet. Each bompus is a felpus. Bompuses are yerpuses. \underline{Felpuses are not fast.} Each felpus is a terpus. Each timpus is fast. Felpuses are quimpuses. Quimpuses are nervous. Each yerpus is not discordant. Each boompus is sunny. Storpuses are wooden. Every zhorpus is brown. Every kerpus is earthy. Kerpuses are rorpuses. \underline{Fae is a felpus}. Fae is a kerpus. \\Question: \\True or false: Fae is fast.\\Answer: \\False
\end{displayquote}
We could see that the question's answer only depends on the fact, where the inference chain is \underline{underlined}. So it's possible that the different fact causes the same queries share different answer. In the PrOntoQA reverse dataset, we harmonized the answers to the same questions in the train dataset. In each sequence, the question could be referred to the form 'A is B?'. We define the OOV dataset as the A and B have never appeared in the train dataset. The accuracy on OOV dataset reflects whether the model learned the rule behind PrOntoQA. These could refer to Fig.~\ref{fig:prontoqa_flowchart}.

Two new versions are involved in the paper, ProsQA and PrOntoQA cloze. The cloze-style version transforms the question `Question: True or false: Fae is fast. Answer: False' into `Question: Fae is \underline{\hbox to 10mm{}} Answer: fast.' The ProsQA version comes from \citep{coconut}, prepares a disturbance options on the result: 
\begin{displayquote}
    Question: Fae is fast or shy? Answer: fast.
\end{displayquote}

We used 500,000 samples for training and 5,000 samples for validation or testing with respect to every PrOntoQA experiment (original, cloze, and reverse). We applied all the data in ProsQA, where there are 18,186 samples for train and 500 for test.

\paragraph{LogicInference}
he LogicInference dataset primarily comprises propositional logic problems and a curated subset of first-order logic formulations. We conducted a two-stage filtering process: initially isolating the first-order logic instances, followed by selecting those containing well-formed yes/no question-answer pairs that are particularly suited for CTP.
\begin{displayquote}
    Fact: \\
    Consider the following premises. exists x15: R15(x15) $\rightarrow$ U1(x15). forall x15: Q15(x15) $\rightarrow$ Q10(x15). forall x15: \textasciitilde P15(x15) or R15(x15). forall x15: P15(x15) or Q15(x15). forall x15: Q(x15). forall x15: Q10(x15) $\rightarrow$ U1(x15). \\
    Question:\\
    Can we infer exists x15: U1(x15) and Q(x15) from them? \\
    Answer: yes
\end{displayquote}

\paragraph{CLUTRR}
CLUTRR is a diagnostic benchmark designed to evaluate the robustness of natural language understanding systems. It tasks models with inferring kinship relations from short stories, requiring both relationship extraction and logical rule deduction. Each story features a complete family structure and requires the model to infer the relationships between any two family members.

\begin{displayquote}
Facts:\\
Stella's husband, Albertus, surprised her with tickets to a football game for their anniversary. Albertus rushed to the hospital to find out that his wife had already given birth to a boy and had named him Pleasant. Frank told a secret to her sister, Blanche. Blanche passed it along to her brother, Pleasant. Pleasant took his Aunt Frank out for her favorite meal. Barnett is Frank's older brother. He has never liked any of her boyfriends. Blanche and her aunt, Frank, went to the deli. They got half a pound of corned beef and two pounds of salami. Gina asked her daughter, Frank, if she had fun at school that day. Frank answered that she and her sister, Frank, had lots of fun together. Albertus went to the game with his sister Frank. Albertus took his daughter Gertie to the park that afternoon to play. Pleasant's wife, Celestia, surprised him on his birthday. He couldn't believe she pulled it off. Florence and her son's wife, Celestia, flew first class to see the concert. \\
    Question: Blanche is who of Stella \\Answer: daughter
\end{displayquote}
\paragraph{LogicAsker}
LogicAsker systematically assesses reasoning by employing atomic skills based on propositional and predicate logic. The LogicAsker dataset features relatively low difficulty and contains few distractors. We sampled 500,000 data for train and 12,000 data for test. An example is:
\begin{displayquote}
    Statement:\\
For all x12, x12 will go running. For all x12, x12 is a police officer. There is at least one x12 for which if x12 were a scientist, then x12 is not a police officer. \\
    Question:\\
Can we infer the following from them? Answer yes or no: There is at least one x12 for which x12 is not a scientist
\\Answer: yes
\end{displayquote}

\paragraph{ReCOGS}
Based on the COGS dataset, ReCOGS is designed for predicting the logical forms of sentences while omitting semantically irrelevant details. It consists of 135,547 train and 3,000 test sequences. THe too short input limits the NTP ability to generalize.

\begin{displayquote}
    Input: \\The cookie was passed to Emma .\\ Output:\\ * cookie ( 32 ) ; Emma ( 22 ) ; pass ( 8 ) AND theme ( 8 , 32 ) AND recipient ( 8 , 22 )
\end{displayquote}

\paragraph{PARARULE Plus}
PARARULE Plus is a deep multi-step reasoning dataset over natural language based on the closed-world assumption. It is derived from the PARARULE dataset and has deeper samples. Similar with the PrOntoQA dataset, it also consists of facts, question and answer. However, it surpasses PrOntoQA in terms of sentence complexity.

However, there is an implicit unreasonable settings in the original dataset, is that all the queries with the answer `true' are end up with the format `A is B?' and the queries with the answer `false' are end up with the format `A is not B?' This causes the transformer learns a shortcut, mapping from existence of `not' in question to the binary answer true or false. From the original settings, both CTP and NTP could easily reach accuracy 1.

We took a deep insight in the generalization rules of PARARULE plus, and rewrote some of them to decouple the answers from the format of queries. We added 4 new rules and redo the same experiments. We use depth-2 dataset for train (500,000 samples) and for test (5,000 samples). 
\begin{displayquote}
    Fact:\\
    The wolf is tired. The wolf is dull. The wolf is rough. The wolf needs the dog. The bear sees the rabbit. \underline{The bear is fierce. The bear is awful.} The dog is kind. The dog is smart. The dog is round. The rabbit is cute. The rabbit is lovely. The rabbit is furry. Kind animals are cute. If something is dull then it visits the dog. If something visits the dog then it is slow. If something is tired and dull then it is rough. If something is cute and lovely then it is adorable. \underline{If something is fierce and awful then it is obese}. If something is rough then it is lazy. All lazy animals are sleepy. If something is cute then it is lovely. All lovely animals are furry. \underline{If something is obese then it is strong.} \underline{All strong animals are heavy}. If something is adorable then it is beautiful. All beautiful animals are small. All slow animals are big. \\
    Question:\\
    The bear is not heavy\\
    Answer: false
\end{displayquote}

\paragraph{RobustLR}
The authors propose RobustLR for diagnose the robustness to logical variations in language models. Compared to PrOntoQA, this dataset is more comprehensive and specific, while also encompassing a variety of different relations. As a consequence, both NTP and LTP face difficulties learning this problem. The LTP's accuracy is stagnated at the random guessing accuracy. The train and test dataset consist of 210,865 and 8,000 samples separately.

\begin{displayquote}
    Statements:\\
    \textcolor[HTML]{42ff8b}{Fiona is white. Dave is blue. Anne is the uncle of Bob.} Charlie is white if Dave is blue. Charlie is white and \textcolor[HTML]{a1808a}{Dave is not quiet if Fiona is white or Anne is the uncle of Bob.} If Fiona is white or Anne is Bob's uncle then Charlie is white and The uncle of Anne is not Gary. If Charlie is white then Anne is big. \textcolor[HTML]{a1808a}{Bob is nice if Dave is not quiet and Anne is the uncle of Bob.} \textcolor[HTML]{ff0088}{Bob is not nice if Anne is the uncle of Bob and Gary is the mother of Harry.} If Dave is blue or Anne is big then Dave is not nice and Bob is nice. If Dave is not quiet and Gary is the mother of Harry then Dave is nice. If Bob is nice or Dave is not nice then Fiona is the aunt of Bob. If Dave is not nice then Bob is not Anne's brother. Bob is Anne's brother if The mother of Harry is Gary. Harry is furry if The brother of Anne is not Bob or Gary is not the uncle of Anne. If Charlie is white and Gary is the mother of Harry then Harry is not furry. Anne is not the wife of Dave if Bob is nice and Anne is Bob's uncle. 
    \\Question:\\ The mother of Harry is not Gary.\\Answer: True
\end{displayquote}
The statement is confusing and we split it into several parts: \textcolor[HTML]{42ff8b}{Facts}, 2-hop \textcolor[HTML]{a1808a}{Inference} and \textcolor[HTML]{ff0088}{contradiction}. 
\paragraph{RuleTaker}
The authors developed the RuleTaker dataset through a systematic transformation of natural language into structured reasoning processes, establishing an emulation framework for soft reasoning. For example, we have following sample like:
\begin{displayquote}
    Statement:\\
    Cow sees mouse. Cow likes tiger. Bear is cold. Cow is big. If X visits bald eagle and X is kind then X is nice.\\
    Question:\\Cow sees bear? \\Answer: False
\end{displayquote}
We use 29,000 samples for training and 1000 for testing. 

\paragraph{SimpleLogic}
Aiming to discover the logic capability in BERT models, especially for its OOD generalization ability, the authors constructed the SimpleLogic dataset, with rule-priority and label-priority. We introduce 192,000 training dataset and 1,0000 testing dataset for this task. The example is attached below:
\begin{displayquote}
    Assumptions:\\
    If messy and reserved, then worrisome. If messy and reserved and tender, then weary. If tender, then friendly. If frightened and worrisome, then tender. If reserved, then tender. If weary, then messy. If lonely and weary and tender, then reserved. If tender, then messy. If worrisome and tender and lonely, then messy. If lonely and frightened and friendly, then messy. If reserved and messy and friendly, then worrisome. If reserved, then frightened. If lonely and friendly and messy, then tender. If frightened, then tender. If lonely, then frightened. If lonely, then worrisome. If messy and friendly, then lonely. If weary, then reserved. If reserved and frightened and weary, then tender. If worrisome and reserved and weary, then frightened. If reserved and friendly, then worrisome. If worrisome, then lonely. If messy and worrisome, then lonely. If frightened, then messy. If lonely, then friendly. If weary, then lonely. \\
    Question: weary worrisome reserved lonely to messy \\
    Answer: true
\end{displayquote}
\paragraph{StepGame}
StepGame is inspired from bAbl-17/19 benchmarks \citep{babi} and to mitigate bAbl's limitations, such as fixed expressions, small number of reasoning hops and the lack of noise for robustness test. Each data instance in the dataset describes a set of spatial relationships among multiple objects and requires the model to deduce the relative position between two specified objects based on the given relational information. Similar to PrOntoQA, we generate 500,000 synthetic training dataset and 5,000 testing dataset. 
\begin{displayquote}
    The object Z is positioned directly above the object K. Object G is above object I and to the right of it, too. N is diagonally to the bottom left of J. A is to the bottom-left of N. K is positioned below and to the right of Y. O is at the lower side of G. Z is to the right of Y. S is placed in the left direction of K. O is directly south east of H. G is to the right of Q. H is placed at the lower right of K. \\ Question: What is the relation of the agent O to the agent G?\\Answer: below
\end{displayquote}
\paragraph{SNLI}
The Stanford Natural Language Inference (SNLI) corpus collects of 570k human-written English sentence pairs for entailment examination. There are 550,152 and 1,000 samples in training and testing dataset. A typical example of SNLI is
\begin{displayquote}
    Text: A man inspects the uniform of a figure in some East Asian country.\\
    Hypothesis: The man is sleeping\\
    Answer: contradiction
\end{displayquote}
\paragraph{Yelp}
The Yelp Dataset is a comprehensive collection of data related to reviews of businesses, and is widely used to predicting positive or negative reviews. We use all the 650,000 sequences for training and 50,000 for testing. The format of reviews like:
\begin{displayquote}
    Text: To keep it short and sweet: Save yourself \$100. Buy a good board game, your alcohol of choice, order a pizza, and invite your friends over. nWhat an incredible disappointment. After seeing the enticing commercials so many times, we decided to give this place a try on a double date. I understand the prices of the play cards and won't dispute them; however, the food was incredibly over-priced, came out COLD (as in, sat on a counter without warmers for a minimum of 30 minutes) and I literally had to ask the bartender if there was any vodka in my drink. It was pure juice. \$38 for three shots that had little-no alcohol in them. (Not to mention, my glass was dirty, and I saw the bartender scoop the glass into the ice basin because she was too lazy to use the sanitary scoop. I know the Food and Beverage Commission would be as disappointed as I was.) The service was terrible. Don't ask for anything from your waiter, as they are a little too busy on their cell phones or conversing amongst themselves. Was it fun to be in an adult-themed arcade? Yes. If you're looking for a good atmosphere to go with friends to play games, I suppose I would advise you give it a shot. I would never recommend their food, customer service, or drinks. Save yourself the money and stay home, or go for a traditional bowling, figure skating, roller-blading, rock climbing, basically any other physically-entertaining themed date instead.\\
    Answer: Negative
    
\end{displayquote}
\paragraph{DBpedia}
The DBpedia dataset is designed to evaluate a model's capability to accurately classify news articles into predefined categories based solely on their titles and concise summaries, thereby testing both the model's comprehension of textual semantics and its ability to perform hierarchical classification tasks. The size of training and testing set are 560,000 and 70,000 separately. 
\begin{displayquote}
    Title: Export-Import Bank of Romania \\ Content:  Exim Bank is The Export-Import Bank of Romania based in Bucharest. \\Answer: 0
\end{displayquote}
\subsection{Experimental Framework and Implementation Details}
\label{sec:experiment_detail}
This section provides a detailed description on the experimental implementations.  
\subsubsection{Anchor function}
\label{sec:anchorfunction_appendix}
\begin{figure}[!h]
    \centering
    \includegraphics[width=0.7\linewidth]{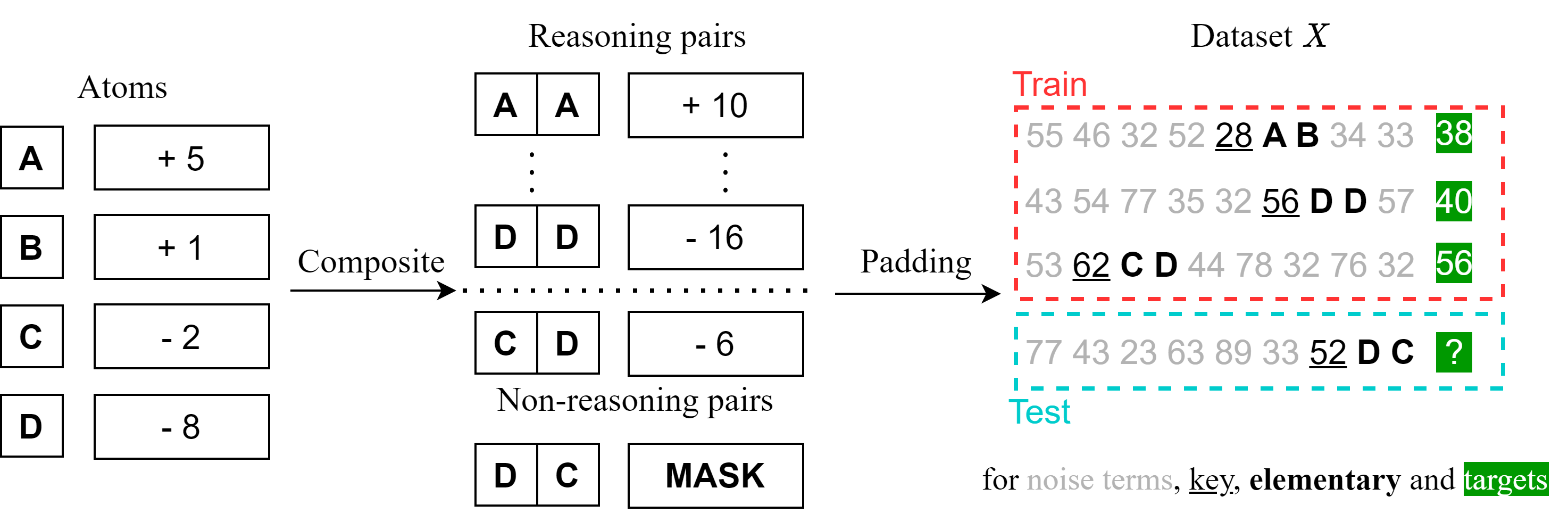}
    \caption{The illustration of anchor function data generation. Anchor function is a simple linear mapping but we add the non-reasoning compositional anchor $(C,D)$ in the training set. We focus on the model preference on its symmetric pair $(D,C)$, which excluded from training. If the prediction on $(D,C)$ match the atom composition rule, it's called reasoning solution, 
 otherwise non-reasoning solution.}
    \label{fig:anchorfunction_flowchart}
\end{figure}
\paragraph{Definition of anchor function}
Consider a function $f(\mathbf{x}): \mathbb R^{s\times d} \to \mathbb R^C$, where $s$ represents for sequence length while $C$ for vocabulary size. The input $X$ consists of two parts: anchor set $\mathcal A = \{A, B, C, D\}$ and the definition domain of function $f$, $\mathcal D = \{20, \ldots, 100\}$. The function is defined as:
\begin{align}
    f(\ldots, x_{i}, a, x_{i+2}, \ldots) &:= a(x_{i}),\quad \text{while}\ \  a\in \mathcal{A}; x_j \in \mathcal{D} \label{eq:simpleanchor}\\
    f(\ldots, x_{i}, a, b, x_{i+3}, \ldots) & = (a,b)(x_i) := b(a(x_{i})),\quad \text{while} \ \ a, b \in \mathcal{A}; x_j \in \mathcal{D} \label{eq:compositionanchor}.
\end{align}
In this work, we set the specific elementary function $f_a, f_b$ as:
\begin{equation*}
    A(x) = x + 5, \quad B(x) = x + 1,\quad C(x) = x - 2,\quad D(x) = x - 8
\end{equation*}
The anchor function operates solely on the position preceding of anchor $a$, which is denoted as key item, and is independent of the input at other positions. However, the pairs $(C,D)$ and $(D,C)$ have their own specific pattern modes. The pair $(C,D)$ is defined as the non-reasoning solution and available in training dataset,
    \begin{equation*}
        (C,D)(x) := x - 6
    \end{equation*}
    while $(D,C)$ is defined as the reasoning solution $(D,C)(x) = x - 10$ and still masked from training. 
\paragraph{Data Generation}
Since we have fixed the anchor set $\mathcal{A}$, then for composition task shown in Eq.~\eqref{eq:compositionanchor}, 16 anchor pairs exist in total. We generate 900,000 samples in total and partition it into training and testing subsets with a 9:1 ratio. Each anchor pair $(a,b)$ shares the equal number of samples. Then we generate the dataset $X$: The position of anchor and key are randomly selected in the fixed-length sequence, and the other positions are filled with random number from $\mathcal{D}$. The last token is replaced by the function solution of the sequence, i.e.
\begin{equation}
    X = \{x_i\in \mathcal{D}, a,b \in \mathcal A\Big| [x_1, \ldots, x_i, a, b, \ldots, x_n, (a,b)(x_i)]\}.
\end{equation}
An example is $[56, 74, 65, D, C, 89, 84, 41, 34, 55]$, where we have $f_{DC}(65) = 65 - 10 = 55$.

% \subsection{Experimental Setup}
We use vallina transformers with AdamW optimizer, learning rate is set as 2e-5 with linear warmup scheduler and the weight decay is set as 0, since we tend to exclude the effect of regularization methods. The batch size is set to 2000.

\paragraph{Model architecture}
In Fig.~\ref{fig:NTP Training process}, the transformer is set from 3 layers and 8 layers and 2 heads to 4 heads, with 400 hidden state dimension and 64 dimension for $Q, K, V$. We choose ReLU as the activation in $\mathrm{MLP}$ blocks with dimension 1200. We apply the kaiming normal initialization on the transformer layers. In the experiments we noticed these hyperparameters exhibit minimal impact on the experimental outcomes.

% \paragraph{Training Method}
% Here we denote $F$ as the transformer and rewrite the NTP and CTP loss functions for further analysis. The crossentropy loss of NTP and CTP could be written as:
% \begin{align}
%     \mathcal L_N &= -\frac{1}{NS}\sum_{j=1}^N\sum_{s=1}^{S}\sum_{c=1}^{C}\text 1\{x_{s+1}^j=c\}\log(\sigma(\hat x_{s}^j)_c)\\
%     \mathcal L_L &= -\frac{1}{N}\sum_{j=1}^N\sum_{c=1}^{C}\text 1\{x^j=c\}\log(
% \sigma(\hat x^j)_c).\\
% \mathcal L_N &= \frac{1}{S} \mathcal L_L + \mathcal L_{noise}
% \end{align}
% where $\hat x^j = F(x^j,\theta) \in \mathbb R^{s\times C}$. 
% With the gradient of CTP and NTP loss, we see:
% \begin{align}
% \nabla_{\theta}{\mathcal L_L} &= -\frac{1}{N}\sum_{j=1}^N\sum_{c=1}^{C}(1\{y^j=c\} - \sigma{(\hat x^j)_c})\pp{ f_{S,c}}{\theta} \\
% &=-\sum_{c=1}^{C}\pp{ f_{S,c}}{\theta}\frac{1}{N}\sum_{j=1}^N(1\{y^j=c\} - \sigma{(\hat x^j)_c})\\
% &\sim-\sum_{c=1}^{C}\pp{ f_{S,c}}{\theta}(\frac{1}{C}-\overline{\sigma{(\hat{x}^j)_c}})
% \end{align}

% About uniform loss $\mathcal{L}_{uni}$, it could be interpreted as a motivate for transformer learning constant mapping, from $x$ to $(1/C, \ldots, 1/C)$, which shares the same regularization idea with the SAM or data argumentation methods, which contribute to more flat and well-generalized models.

\subsubsection{Reasoning tasks}
For reasoning tasks, we use vallina GPT-2 model with 12 layers and 12 heads, embedding dimension is set as 768. We forbid dropout in the residual, embedding and attention branch, to avoid effect of regularization methods. We set the learning rate is 5e-5 with linear warmup scheduler. 

To construct the dataset for CTP training, we equipped the answer with a separation mark, use the PrOntoQA for example, we turn the sequence into:

\begin{displayquote}
    Gwompuses are zhorpuses. Every chorpus is transparent. Each gerpus is a boompus. Bompuses are sweet. Each bompus is a felpus. Bompuses are yerpuses. Felpuses are not fast. Each felpus is a terpus. Each timpus is fast. Felpuses are quimpuses. Every zhorpus is brown. Every kerpus is earthy. Kerpuses are rorpuses. Fae is a felpus. Fae is a kerpus. Question: True or false: Fae is fast. \textbf{[SEP]} False \textbf{[SEP]} 
\end{displayquote}
Like SFT, the loss of CTP is only calculated tokens between the \textbf{[SEP]} symbols. 

\subsubsection{Addition task}
The addition task is designed to show the robustness of NTP training. We borrow the reverse addition settings, like $314 + 518 = 832$, changed into $413 + 815 = 238$. Given the pure addition doesn't contain any noise in the corpus, We intentionally introduce noise tokens into the dataset. The reconstructed sequence is, for example, 
\begin{displayquote}
7, 9, 1, 1, [SEP\_R], 5, 5, 4, 0, +, 3, 5, 4, 0, [SEP\_R], 4,
    \textbf{[SEP]} 8, 0, 9, 0 \textbf{[SEP]} 
\end{displayquote}
The symbol [SEP\_R] is used to remind the model of the start and end in equation, and the first \textbf{[SEP]} could be regarded as the equal symbol `='. Outside the symbol [SEP\_R], we add 5 noise terms to help simulate the noise in anchor function. The training configurations follow the anchor function with a 8 layers 4 heads prenorm model.
\end{document}